\pdfoutput=1

\documentclass[11pt]{article}

\usepackage{acl}

\usepackage{times}
\usepackage{float}
\usepackage{enumitem}
\usepackage{url}
\usepackage{multirow} 
\usepackage{graphicx}
\usepackage{booktabs}
\usepackage{latexsym}
\usepackage{subfig}
\usepackage{amssymb}
\usepackage{natbib}
\usepackage{amsmath}
\usepackage{algorithm}
\usepackage{algpseudocode}
\usepackage[T1]{fontenc}

\usepackage[utf8]{inputenc}

\usepackage{microtype}

\usepackage{array}

\title{Beyond Single-Event Extraction: 
Towards \\  Efficient 
Document-Level Multi-Event Argument Extraction}

\author{Wanlong Liu\textsuperscript{\rm 1},
    Li Zhou\textsuperscript{\rm 1},
    Dingyi Zeng\textsuperscript{\rm 1},
    Yichen Xiao\textsuperscript{\rm 1},\\  \textbf{Shaohuan Cheng}\textsuperscript{\rm 1},
    \textbf{Chen Zhang}\textsuperscript{\rm 2},
    \textbf{Grandee Lee}\textsuperscript{\rm 3},
    \textbf{Malu Zhang}\textsuperscript{\rm 1}\thanks{~~Corresponding author}, 
    \textbf{Wenyu Chen}\textsuperscript{\rm 1}
    \\
        {\textsuperscript{\rm 1}{University of Electronic Science and Technology of China}} \\ {\textsuperscript{\rm 2}{	National University of Singapore}} \\ {\textsuperscript{\rm 3}{Singapore University of Social Sciences}} \\liuwanlong@std.uestc.edu.cn, maluzhang@uestc.edu.cn,
        cwy@uestc.edu.cn}

\begin{document}
\maketitle
\begin{abstract}
Recent mainstream event argument extraction methods process each event in isolation, resulting in inefficient inference and ignoring the correlations among multiple events.  
To address these limitations, here we propose a multiple-event argument extraction model DEEIA (\textit{\textbf{D}ependency-guided \textbf{E}ncoding and \textbf{E}vent-specific \textbf{I}nformation \textbf{A}ggregation}), capable of extracting arguments from all events within a document simultaneously. 
The proposed DEEIA model employs a multi-event prompt mechanism, comprising DE and EIA modules. 
The DE module is designed to improve the correlation between prompts and their corresponding event contexts, whereas the EIA module provides event-specific information to improve contextual understanding.
Extensive experiments show that our method achieves new state-of-the-art performance on {four} public datasets (RAMS, WikiEvents, MLEE, and {ACE05}), while significantly saving the inference time compared to the baselines.
Further analyses demonstrate the effectiveness of the proposed modules. Our implementation is available at \url{https://github.com/LWL-cpu/DEEIA}.
\end{abstract}




\section{Introduction}
\begin{figure}[t!]
    \centering
    \includegraphics[width=1.0\linewidth]{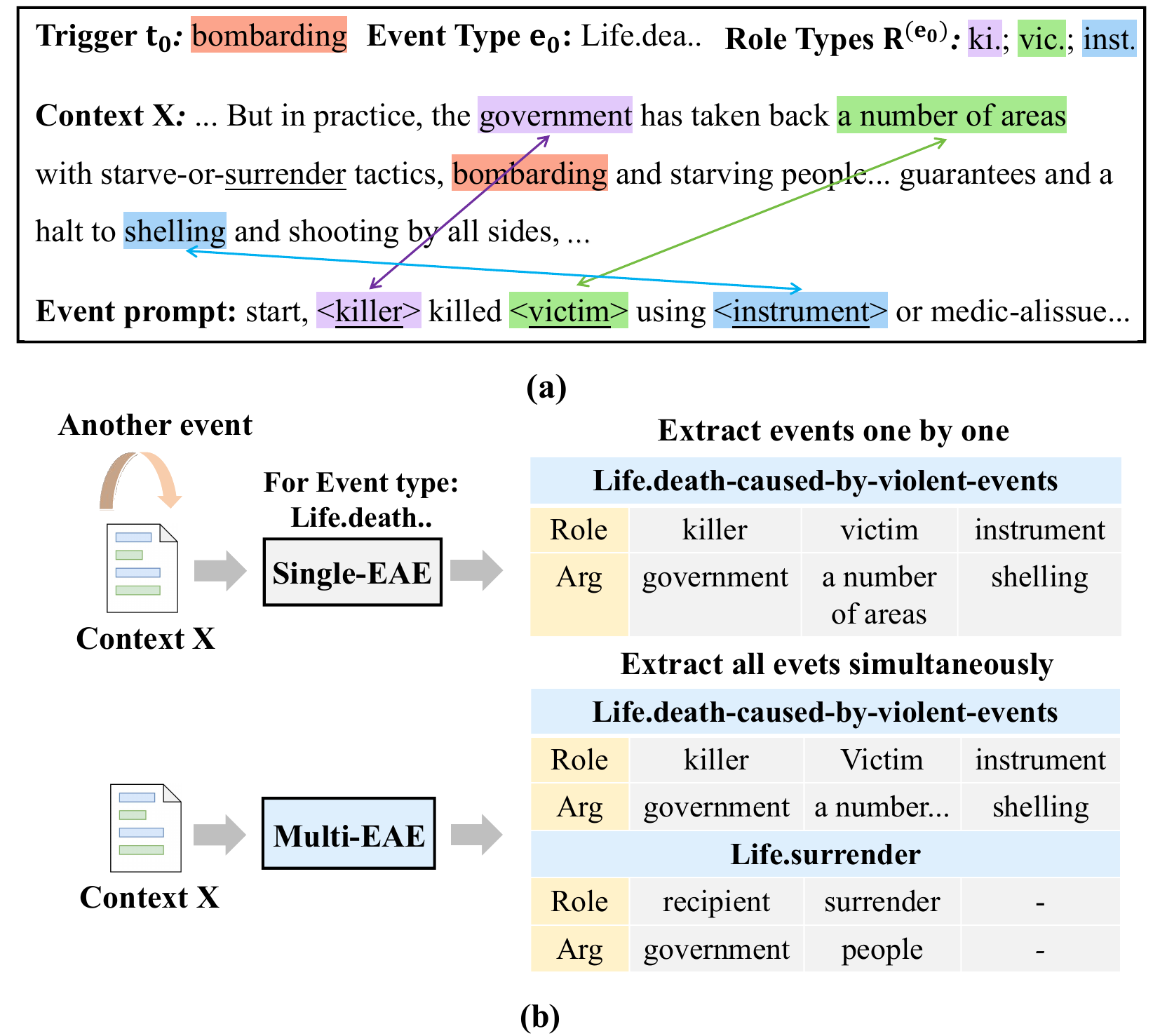}
    
    \caption{
     Subfigure (a) explains the prompt-based EAE task with one of the events in context $X$. The prompt is manually designed for the specified event type, with mentions of roles as \textbf{slots}, such as $\left< killer \right> $. (b) shows the difference between traditional Single-EAE method and our Multi-EAE method, the latter is more difficult.}
    \label{fig:fig1}
\end{figure}

Document-level event argument extraction (EAE) is a key process within Information Extraction~\cite{ hobbs2010information, grishman2015information, xia2022metatkg}, focused on identifying event-related arguments and their respective roles in document-level texts. Recently, the leading-edge methods for this task delve into prompt-based techniques~\cite{ma2022prompt, hsu2023ampere}, due to their great generalizability and competitive
performance.
Figure~\ref{fig:fig1} (a) presents an example demonstrating EAE utilizing a prompt-based approach.
Regarding the  event $e_0$ triggered by ``bombarding'', the approach defines a corresponding event prompt and identifies  arguments: ``government'' as the \textit{killer}, ``a number of areas'' as the \textit{victim}, and ``shelling'' as the \textit{instrument}.

Mainstream EAE works~\cite{zhou2024eace,liu2023enhancing, ren2023retrieve, liu-etal-2023-document} can only process one event at a time. 
When encountering documents containing multiple events, the limitations of these single-event argument extraction (Single-EAE) methods become evident.  (1) As shown in Figure~\ref{fig:fig1} (b), for a document containing multiple events,  Single-EAE methods  have to perform numerous iterations to extract event arguments for all events, which process the same document text repeatedly, leading to inefficient extraction. (2)  Single-EAE methods fail to capture the beneficial event correlations among multiple events~\cite{he2023revisiting, zeng2022ea}.  Figure~\ref{fig:fig1} (b) illustrates the argument overlapping phenomenon which reflects the semantic correlations among events. The event \texttt{Life.death} and event \texttt{Life.surrender} share the argument ``government'' and there exists a strong event correlation between these two events. However, Single-EAE methods cannot utilize such correlations.   


To tackle these limitations, this paper proposes a DEEIA (\textit{\textbf{D}ependency-guided \textbf{E}ncoding and \textbf{E}vent-specific \textbf{I}nformation \textbf{A}ggregation}) model, a multiple-event argument extraction (Multi-EAE) method capable of simultaneously extracting arguments for all events within the document. 
We construct our DEEIA model based on the state-of-the-art (SOTA) prompt-based Single-EAE model PAIE~\cite{ma2022prompt} and introduce a multi-event prompt mechanism to enable extracting arguments from multiple events simultaneously.

However, our Multi-EAE model faces the challenge of handling more complex information as it needs to simultaneously process different triggers, arguments roles, and prompts from multiple events. This requires the model with enhanced information extraction capabilities.
Therefore, we design a Dependency-guided Encoding (DE) module to guide the model in correlating the various prompts with their respective event contexts.
Furthermore, we propose an Event-specific Information Aggregation (EIA) module to provide event-specific contextual information for a better context understanding.  

Figure~\ref{fig:fig:time} demonstrates that with the increase in the number of events within a document, the efficiency advantage of our DEEIA model becomes increasingly apparent. 
The performance surpassing Single-EAE baselines also demonstrates that our model effectively captures event correlations.
The contributions of this paper are summarized as follows:

\begin{figure}[tbp]
    \centering
    \includegraphics[width=1.0\linewidth]{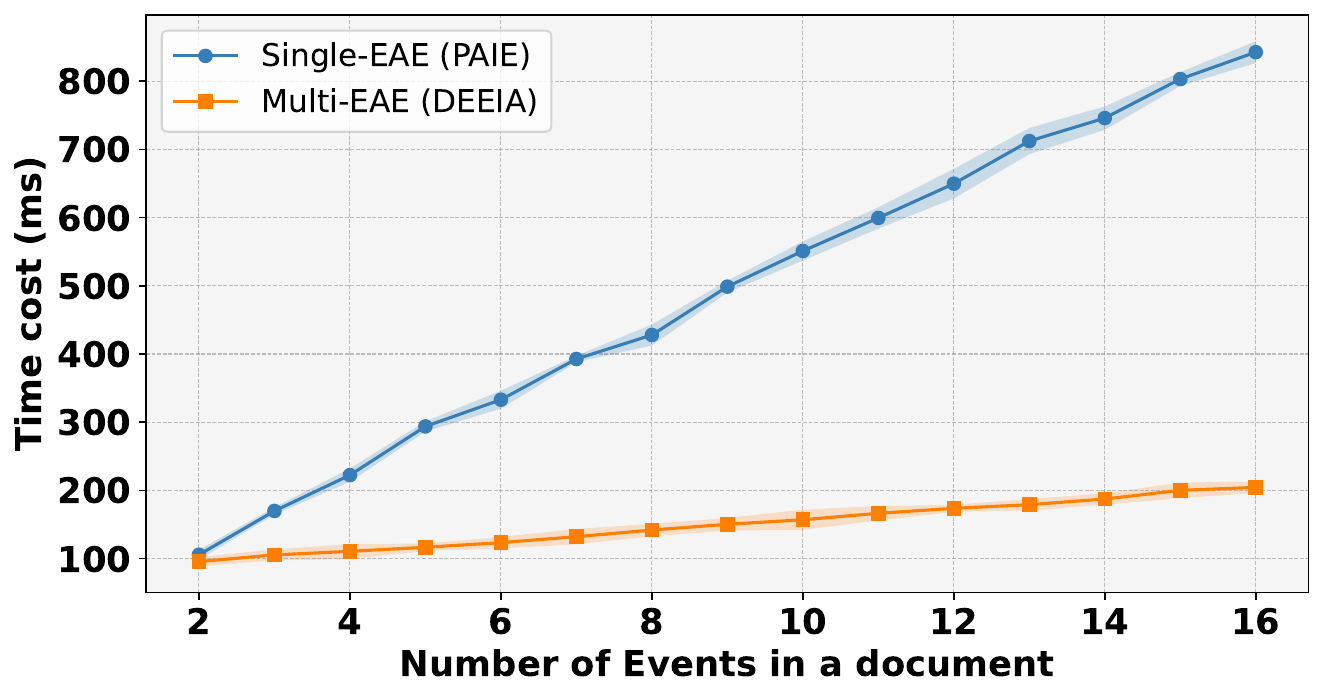}
    
    \caption{ 
     We select document samples containing
     different numbers of events and calculate the inference time on one sample  for a Single-EAE method PAIE~\cite{ma2022prompt} and our Multi-EAE model DEEIA. The results are averaged on 100 repeated experiments. With the increase of event numbers within a document, the efficiency advantage of our Multi-EAE model becomes increasingly apparent.}
    \label{fig:fig:time}
\end{figure}  

\begin{itemize} 
\setlength{\itemsep}{1pt}
\setlength{\parsep}{1pt}
\setlength{\parskip}{0pt}
\item  We propose a multi-event argument extraction (Multi-EAE) method, capable of extracting the arguments of multiple events simultaneously, with the aim of improving the efficiency and performance of the EAE task.
\item To tackle the challenges of Multi-EAE, we propose a Dependency-guided Encoding (DE) module and an Event-specific Information Aggregation (EIA) module, which provide dependency guidance and event-specific context information, respectively. 

\item Extensive experiments demonstrate that the proposed DEEIA model outperforms major benchmarks in terms of both performance and inference time. We provide comprehensive ablation studies and analyses.

\end{itemize}

\begin{figure*}[htbp]
    \centering
    \includegraphics[width=1.0\linewidth]{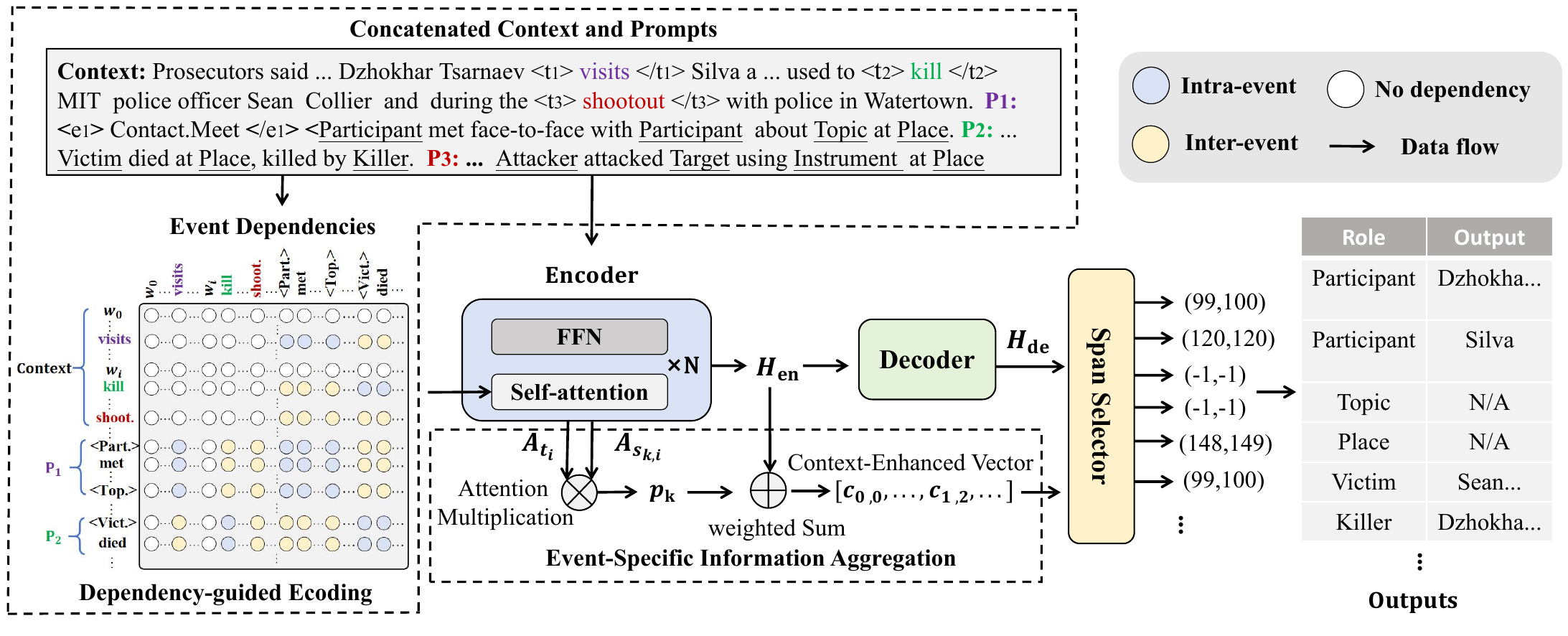}
    
    \caption{The architecture of the proposed DEEIA model. For an input document, $\text{P}_1$, $\text{P}_2$, and $\text{P}_3$ represent simplified prompts. }
    \label{fig:architecture of model}
\end{figure*}  

\section{Related Work}
Recently, there has been an increasing interest in the task of document-level Event Argument Extraction~\cite{wangetal2022query, liu2023document, yang2023amr}, a crucial component within the domain of  Event Extraction ~\cite{ren2022clio, yang2021document}.  
Current methods for document-level EAE can be classified into four main categories: (1) Span-based methods, which identify candidate spans and subsequently predict their roles~\cite{ zhang2020two, yang2023amr, liuetal2017exploiting, zhang2020two, liu2023enhancing}. (2) Generation-based methods~\cite{lietal2021document, du2021template, weietal2021trigger, huang2023simple} utilizing generative PLMs, such as BART~\cite{lewis2019bart}, to sequentially produce all arguments for the designated event. (3) Prompt-based methods~\cite{ma2022prompt, he2023revisiting, nguyen2023contextualized, wang2024degap}, which use slotted prompts and leverage a generative slot-filling approach for argument extraction. (4) Large language model methods. 
Recently, some work~\cite{zhang2024ultra, zhou2023heuristics} has attempted to explore to utilize large language models for EAE tasks, but the performance falls short of expectations. And the time and cost of inference are relatively high. Among them, prompt-based methods have been demonstrated superior generalizability and competitive
performance~\cite{ hsu2023ampere}.   




However, these EAE methods are Single-EAE methods, which can only process one event at a time.  
Recently, prompt-based EAE method, TabEAE~\cite{he2023revisiting} aims to capture event co-occurrence~\cite{zeng2022ea} and trains the model on multi-event scheme.
However, TabEAE requires separately processing prompts for different events, which is highly time-consuming.  
In this paper, we propose to extract event arguments concurrently, which significantly improves the efficiency of the EAE task in multi-event documents.



\section{Methodology}
In this section, we first provide a formal definition of multi-EAE task.
Given an instance $\left( X, \{e_i\}^K_{i=1}, \{t_i\}^K_{i=1},  \{R^{\left( e_i \right)} \}^K_{i=1} \right) $, where $X=\left( w_0, w_2, …, w_{N-1} \right) $ represents the document text with $N$ words, $K$ is the number of target events,  and $e_i$ is the type of the $i$-th event. $t_i\subseteq X$ is the trigger word of the $i$-th event, and $ R^{\left( e_i \right)}$ indicates the set of roles associated with event $e_i$.
The task aims to extract a set of span $\mathcal{S}_i$ for each event $e_i$, which satisfies $\forall a^{\left( r \right)}\in \mathcal{S}_i,(a^{\left( r \right)}\subseteq X)\land (r\in R^{\left( e_i \right)})$. 
Most previous EAE methods are designed as Single-EAE methods, applicable when $K=1$.

We then provide a detailed description of our DEEIA model, as shown in Figure~\ref{fig:architecture of model}. We first propose a multi-event prompt mechanism to enable extracting arguments
from multiple events simultaneously (\S~\ref{Sec:Multi-event Prompt Mechanism}). Then we propose a Dependency-guided Encoding module (\S~\ref{Sec:Dependency-guided Encoding module}) and an Event-specific Information Aggregation module (\S~\ref{Sec:Event-specific Information Aggregation}) to tackle the complexity and challenge of simultaneously
extracting arguments of multiple events.

\subsection{Multi-event Prompt Mechanism}
\label{Sec:Multi-event Prompt Mechanism}

Multi-event document instances involve the prompts of multiple events.  Therefore, we propose a multi-event prompt mechanism to enable extracting arguments
from multiple events simultaneously. We first enhance the input text and event prompts and then concatenate them as the final input. 

\textbf{Preprocessed Text.} \quad
Given an input text with a set
of event triggers, each trigger is initially annotated with a unique pair of markers ($\left<t_i\right>, \left</t_i\right>$), where $i$ counts
the order of occurrence. 
Then we tokenize the marked text into $X$, where $t_i$
is the $i$-th trigger:
\begin{equation}
 \begin{array}{c}
    \hat{X}=  w_0 \left<t_0\right> t_0 \left</t_0\right>  ... \left<t_i\right>  t_i  \left</t_i\right>  ...  w_N.
\end{array}
\end{equation}

\textbf{Prompt Enhancement.}
\label{Prompt Enhancement.}
We concatenate the prompt of each event and obtain the final prompt $P$.\footnote{If multiple events are of the same type, we retain only one prompt and use it to predict arguments of all such events.}
For each event prompt $P_i$, we utilize the event-schema prompts proposed by PAIE~\cite{ma2022prompt}. We append the event type to the start of each corresponding prompt. This strategy helps the model distinguish different event prompts and integrates event type information, enriching the EAE process with more comprehensive information.
Specifically, we wrap each event type with a unique pair of markers ($\left<e_i\right>, \left</e_i\right>$) and obtain the final $P$ as follows:

\begin{equation}
\setlength\abovedisplayskip{3pt plus 3pt }
\setlength\belowdisplayskip{3pt plus 3pt }
 \begin{array}{c}
    P= \left<e_0\right> w_0^{e_0} ... \left</e_0\right> P_0 ... \left<e_i\right> 
    w_0^{e_i}... \left</e_i\right> P_i, 
\end{array}
\end{equation}
where $P_i$ is the prompt of the $i$-th event and $w_0^{e_i}$ is the first token of event type $e_i$.

Then we concatenate $\hat{X}$ and $P$ and put them into our dependency-guided encoder (in \S~\ref{subsub:Dependency-guided Structural Encoder}). For input sequences that exceed the maximum length\footnote{The length of event prompts is much shorter than that of the text, and only a very small number of docs in
Wikievents and MLEE datasets exceed the length limit.}, we employ a dynamic window~\cite{zhou2021document} technique to solve the problem, where the detailed algorithm is shown in .

\subsection{Dependency-guided Encoding Module}
\label{Sec:Dependency-guided Encoding module}
\subsubsection{Event Dependency Definition}
\label{sec:dependency defination}
Due to the fact that different events are associated with distinct trigger words, argument roles, and prompts, our model faces the challenge of information complexity~\cite{bagga1997analyzing, li2023intra} when processing multiple events simultaneously.
Therefore, we propose to guide the model to associate the multi-event prompts with their corresponding event contexts with pre-defined dependencies.
Considering that arguments can exhibit inter-event and intra-event relations within a context, we define the following two types of dependencies among triggers and prompts (including argument slots), which help to solve the information complexity problem of multiple events.

 \textbf{Intra-Event Dependency.} (1) The connection between the trigger and prompt tokens within the same event. (2) The connection between prompt tokens  within the same event prompt. 
 
 \textbf{Inter-Event Dependency.} (1) The connection between the trigger and prompt tokens  of different events. (2) The connection between prompt tokens   of different events.

These two dependencies not only reflect intra and inter-event relations within a multi-event context, but also establish interactions among triggers and argument slots, both within and across events.

Formally, for the input sequence $S=(x_1, x_2, ..., x_n)$ including triggers and prompts from multiple events, we introduce $D=\{dp_{ij}\}$ to represent such two dependencies, where $i,j \in \{0, 1, ..., n\}$ and $dp_{ij} \in \{ \textit{Intra-event, Inter-event, NA} \}$ is a discrete variable denotes the dependency from $x_i$ to $x_j$. $\textit{NA}$ indicates there is no dependency between $x_i$ and $x_j$. Then the  pre-defined event dependencies $D$ will be integrated into the transformer to provide information guidance.

\subsubsection{Dependency-guided Encoder}
\label{subsub:Dependency-guided Structural Encoder}
We improve vanilla self-attention mechanism~\cite{vaswani2017attention} by 
adding a learnable attention bias, which integrates the event dependencies into the transformer.

For the input representation $\mathbf{x}_i \in \mathbb{R}^{d}$, it is first projected into query/key/value vector:
 $\mathbf{q}_i = \mathbf{x}_i\mathbf{W}_Q, \mathbf{k}_i = \mathbf{x}_i\mathbf{W}_K,\textbf{v}_i = \mathbf{x}_i\mathbf{W}_V$.
Then a learnable bias is added to the vanilla self-attention mechanism, which helps the model perceive dependency relations among multiple events. The attention score $ a_{ij}$ is produced as follows:
\begin{equation}
   a_{ij} =  \frac{{\mathbf{q}_i} \mathbf{k}^T_j}{\sqrt{d_k}} + \gamma \cdot bias_{ij},
\end{equation}
where $bias_{ij}$ is the learnable attention bias for the attention between tokens $x_i$ and $x_j$. $\gamma$ is a hyperparameter for adjusting the influence of the bias. $d_k$ is the the hidden dimension of each attention head. Specifically, the attention bias depends on the dependency $dp_{ij}$ and the context information, with specific parameters trained and then utilized in a compositional manner~\cite{xu2021entity}.  We design the attention bias $bias_{ij}$ as follows:
\begin{equation}
bias_{ij} = 
\begin{cases} 
0, & \text{if } dp_{ij} \ is \ \textit{NA}  \\
\frac{{\mathbf{q}_i} \mathbf{W}_{dp_{ij}} \mathbf{k}^T_j + b_{ dp_{ij}}}{\sqrt{d_k}}, &  \text{otherwise}

\end{cases}
\label{eq4}
\end{equation}
where $\mathbf{W}_{dp_{ij}} \in \mathbb{R}^{d_{k} \times 1 \times d_{k}}$ and $b_{ dp_{ij}}$  are the trainable parameters corresponding to dependency $dp_{ij}$. 
The remaining operations are the same as transformer mechanism. Then we apply this mechanism to each layer of the encoder. Note that we do not apply this mechanism to the decoder layers, and a detailed analysis is provided in Appendix~\ref{sec: Architecture Variants}.
\subsection{Event-specific Information Aggregation}
\label{Sec:Event-specific Information Aggregation}
In this section,
we design an event-specific information aggregation (EIA) module which adaptively aggregates useful information for specific events. 
{We hope the model can make use of the context information relevant to the specific event and the argument when extracting an argument. Therefore, we consider using triggers (representing the event) and slots (representing the argument) to measure the relevance between the target argument and context (prompt) information. }

Specifically, we utilize the attention heads of argument slots and their associated triggers, derived from the pre-trained transformer encoder, to calculate the attention product for the input sequence tokens (including both context and prompt).  {The dot product of attention is designed to measure the degree of association between the current event's argument and every token of input context. }

We adopt an encoder-decoder architecture. The encoder is employed to encode the input text, while the decoder is tasked with deriving the event-oriented context and context-oriented prompt
representation $\mathbf{H}_\mathrm{de}$:

\begin{equation}
    \begin{array}{c}
[\mathbf{A}; \mathbf{H}_\mathrm{en}] = \mathrm{Encoder}_s(S), \vspace{1.0ex} \\
\mathbf{H}_\mathrm{de} = \mathrm{Decoder} (\mathbf{H}_\mathrm{en}), \vspace{1.0ex} 
 \end{array}
\end{equation}
where the $\mathrm{Encoder}_s$ is the dependency-guided encoder  and the $\mathrm{Decoder}$ is a transformer-based decoder.
$S$ is the input of the concatenation of context $X$ and prompt $P$, and $\mathbf{A} \in \mathbb{R}^{H \times l_s \times l_s}$ is the multi-head attention matrix and $\mathbf{H}_\mathrm{en}, \mathbf{H}_\mathrm{de}  \in \mathbb{R}^{l_s \times d}$. $H$ is the attention head numbers and $l_s$ is the length of input sequence $S$.

For the $k$-th argument slot $s_{k,i}$ to be predicted in the $i$-th event, we first get the contextual attention vectors $\textbf{A}_{t_i} \in \mathbb{R}^{l_s}$ and $\textbf{A}_{s_{k,i}} \in \mathbb{R}^{l_s}$ from $\mathbf{A}$,  corresponding to the trigger $t_i$ and the slot in the prompt respectively.  These vectors are obtained by averaging across all attention heads and associated subtokens\footnote{We only take the start token $\left<t_i\right>$ to represent the trigger.}.
Then for the argument slot $s_{k,i}$, we obtain the context-enhanced vector $\textbf{c}_{k,i} \in \mathbb{R}^{d}$  which adaptively aggregates useful 
context and prompt information for argument extraction. 

\begin{equation}
\label{eq6}
\setlength\abovedisplayskip{3pt plus 3pt minus 7pt}
\setlength\belowdisplayskip{3pt plus 3pt minus 7pt}
    \begin{array}{c}
\textbf{p}_k=\mathrm{softmax}(\textbf{A}_{t_i}\cdot \textbf{A}_{s_{k,i}}\;)\vspace{1.0ex}, \\
    \textbf{c}_{k,i}={\mathbf{H}_\mathrm{en}}^T \ \textbf{p}_k,
    \end{array}
\end{equation}%
where $\textbf{p}_k \in \mathbb{R}^{l_s}$ is the computed attention weight vector for argument slot $s_{k,i}$.  Then $\textbf{c}_{k,i}$ is subsequently incorporated into the decoder output $\mathbf{h}_{s_{k,i}} \in \mathbb{R}^{d}$ of slot $s_{k,i}$ to get $\tilde{\mathbf{h}}_{s_{k,i}} \in \mathbb{R}^{d}$:

\begin{equation} 
\tilde{\mathbf{h}}_{s_{k,i}}=\mathrm{tanh}(\mathbf{W}_{1}[\mathbf{h}_{s_{k,i}}; \textbf{c}_{k,i}]),
\end{equation}
where $\mathbf{W}_{1} \in \mathbb{R}^{2d \times d}$ is learnable parameter.
\subsection{Span Selection}
After obtaining the final representation $\tilde{\mathbf{h}}_{s_{k,i}}$ for each slot within each event, we follow~\cite{ma2022prompt} and transform each of them into a set of span selector \{$\Phi_{s_{k,i}}^\mathrm{start}, \Phi_{s_{k,i}}^\mathrm{end}$ \}:

\begin{equation} 
\begin{array}{c}
\Phi_{s_{k,i}}^\mathrm{start} = \tilde{\mathbf{h}}_{s_{k,i}} \circ \mathbf{w}_\mathrm{start}, \vspace{1.0ex} \\
\Phi_{s_{k,i}}^\mathrm{end} = \tilde{\mathbf{h}}_{s_{k,i}} \circ \mathbf{w}_\mathrm{end},
\end{array}
\end{equation}
where $\mathbf{w}_\mathrm{start}, \mathbf{w}_\mathrm{end} \in \mathbb{R}^{d}$ are  learnable parameters and $\circ$ represents element-wise multiplication. Then $\Phi_{s_{k,i}}^\mathrm{start}$ and $\Phi_{s_{k,i}}^\mathrm{end}$ determine the start and end positions of slot $s_{k,i}$ in the original text:

\begin{equation} 
\setlength\abovedisplayskip{3pt plus 3pt minus 7pt}
\setlength\belowdisplayskip{3pt plus 3pt minus 7pt}
\begin{array}{c}
\text{logit}^\mathrm{start}_{k,i} = \mathrm{softmax}(\mathbf{H}_\mathrm{de} \Phi_{s_{k,i}}^\mathrm{start}) \in \mathbb{R}^{l_s} \vspace{1.0ex}, \\
\text{logit}^\mathrm{end}_{k,i} = \mathrm{softmax}(\mathbf{H}_\mathrm{de}\Phi_{s_{k,i}}^\mathrm{end}) \in \mathbb{R}^{l_s} \vspace{1.0ex}, \\

\text{score}_{k,i}(m,n) = \text{logit}^\mathrm{start}_{k,i}(m) + \text{logit}^\mathrm{end}_{k,i}(n) \vspace{1.0ex}, \\

(\hat{s}_{k,i}, \hat{e}_{k,i}) = \underset{(m,n) \in C}{\mathrm{arg\,max}}\ \text{score}_{k,i}(m, n),

\end{array}
\end{equation}
where $(\hat{s}_{k,i}, \hat{e}_{k,i})$ is the predicted argument span.  $C = \left\{ (m, n) \mid (m, n) \in {l_s}^2, 0 < n - m \leq l \right\} \cup \{(0, 0)\} 
$ contains all spans not exceeding the threshold \( l \), along with the empty span \( (0, 0) \). 

Following~\cite{ma2022prompt}, we utilize Bipartite Matching Loss~\cite{carion2020end}, which provides further consideration for the assignment of golden argument spans during training. 
\begin{equation}
\setlength\abovedisplayskip{3pt plus 3pt minus 7pt}
\setlength\belowdisplayskip{3pt plus 3pt minus 7pt}
\begin{array}{c}
\mathcal{L} = -\sum\limits_{i=1}^{\mathcal{K}} \sum\limits_{\substack{(\hat{s}_{k,i}, \hat{e}_{k,i})  \in \delta(\mathcal{S}_i)}} 
[\log \text{logit}^\mathrm{start}_{k,i}(\hat{s}_{k,i})  \vspace{1.0ex} \\
+ \log \text{logit}^\mathrm{end}_{k,i}(\hat{e}_{k,i})],
\end{array}
\end{equation}
where $\mathcal{K}$ is the number of target events and $i$ represents the $i$-th event. $ \delta(\mathcal{S}_i)$ denotes the optimal assignment~\cite{ma2022prompt} calculated through the Hungarian algorithm~\cite{kuhn1955hungarian}.

\begin{table*}[htbp]
\centering

\small{
\setlength{\tabcolsep}{2.5mm}{
\begin{tabular}{lllllllll}
\hline
\multirow{2}{*}{Scheme} 
& \multirow{2}{*}{Method}  
& \multirow{2}{*}{PLM}
& \multicolumn{2}{c}{RAMS} & \multicolumn{2}{c}{WikiEvents} & \multicolumn{2}{c}{MLEE} \\ \cline{4-9} 
            \multirow{5}{*}{Span-based}                      &   &    & Arg-I & Arg-C & Arg-I & Arg-C & Arg-I & Arg-C \\ \hline
\multirow{5}{*}{single-event}
& ${\text{TSAR}}$~(\citeyear{xu2022two})  & ${\text{BERT-b}}$    & \quad - & 48.1 & 70.8 & 65.5 &  72.3     & 71.3      \\

& ${\text{TSAR}}$~(\citeyear{xu2022two}) & ${\text{RoBERTa-l}}$       & \quad - & 51.2 & 71.1 & 65.8 & 72.6      & 71.5      \\
& ${\text{SCPRG}}$~(\citeyear{liu2023enhancing})  & ${\text{BERT-b}}$  & 53.9*  & 48.9  &70.1*  & 65.8*  & \quad -      & \quad -      \\ 

& ${\text{SCPRG}}$~(\citeyear{liu2023enhancing}) & ${\text{RoBERTa-l}}$   & 56.7*  & 52.3  &71.3*  & \underline{66.4*}  &  \quad -     & \quad -      \\ \hline

\multirow{5}{*}{Generation}  
& ${\text{DocMRC}}$~(\citeyear{liu-etal-2021-machine})  & ${\text{BERT-b}}$    & \quad - & 45.7 & \quad - & 43.3 & \quad -       &  \quad -     \\
\multirow{5}{*}{single-event }  

& ${\text{EEQA}}$~(\citeyear{du-cardie-2020-event})   & ${\text{BART-l}}$    & 48.7 & 46.7 & 56.9 & 54.5 & 70.3     & 68.7      \\

& ${\text{FEAE}}$~(\citeyear{weietal2021trigger})   & ${\text{BERT-b}}$    & 53.5 & 47.4 & \quad - & \quad - & \quad -    & \quad -       \\ 
 
& ${\text{BART-Gen}}$~(\citeyear{lietal2021document}) & ${\text{BART-l}}$ & 51.2     & 48.6     & 66.8  & 62.4 & 71.0      & 69.8      \\ 
& ${\text{HRA}}$~(\citeyear{ren2023retrieve})  & ${\text{T5-l}}$ & 54.6     & 48.4     & 69.6 & 63.4 & \quad -     & \quad -      \\ 

\hline

\multirow{3}{*}{Prompt-based}  
& ${\text{RKDE}}$~(\citeyear{hu2023role})   & ${\text{BART-l}}$ & 55.1     & 50.3     & 69.1  & 63.8     &  \quad -     &  \quad -     \\ 
\multirow{3}{*}{single-event}

& ${\text{PAIE}}$~(\citeyear{ma2022prompt})   & ${\text{BART-l}}$ & 56.8     & 52.2     & 70.5  & 65.3     &  72.1*     &  70.8*     \\ 
& ${\text{SPEAE}}$~(\citeyear{nguyen2023contextualized})  & ${\text{BART-l}}$ & \underline{58.0}     & \underline{53.3}     & \textbf{71.9} & 66.1     &  \quad -     &  \quad -     \\
& ${\text{TabEAE}}$~(\citeyear{he2023revisiting})  & ${\text{RoBERTa-l}}$ & 57.0     & 52.5     & 70.8  & 65.4     &  71.9     &  71.0       \\

\hline
\multirow{2}{*}{Prompt-based}
& ${\text{PAIE}}$-$\text{multi}$    & ${\text{BART-l}}$  & 55.9 & 50.9 & 67.2 & 61.7 &  71.3     &  69.5     \\ 

\multirow{2}{*}{multi-event}

& $\text{TabEAE}$-$\text{multi}$   & ${\text{RoBERTa-l}}$  & 56.7 & 51.8 & 71.1 & 66.0 &  \underline{75.1}     &  \underline{74.2}     \\ 
& $\text{DEEIA(Ours)}$    & ${\text{RoBERTa-l}}$
&\textbf{58.0} & \textbf{53.4} & \underline{71.8} & \textbf{67.0}  &  \textbf{75.2}     &  \textbf{74.3}             \\ \hline
\end{tabular}
}
\caption{
Comparison of performance on RAMS, WikiEvents and MLEE test set. * means we rerun their code based on their experimental settings. 
\textbf{Bold} and \underline{underline} indicate the best and second-best experimental results. 
}

\label{tab:main_results}
}
\end{table*}

\section{Experiments}
\subsection{Experimental Setup}
\noindent \textbf{Datasets} \quad
We evaluate our model on three document-level EAE datasets, including RAMS~\cite{ebner2020multi},WikiEvents~\cite{lietal2021document} and MLEE~\cite{10.1093/bioinformatics/bts407}. {Moreover, we also extend our evaluation of the model to the sentence-level ACE05 dataset~\cite{doddington2004automatic}, as it includes a significant number of instances with multiple events.} The detailed dataset description and statistics are shown in Appendix~\ref{appendix:data_statictics}.

\noindent \textbf{Evaluation Metrics} \quad
Following previous works~\cite{ma2022prompt, he2023revisiting}, we evaluate performance using two metrics: (1) strict argument identification F1 (Arg-I), where a predicted event argument is considered correct if its boundaries match those of any corresponding golden arguments. (2) Strict argument classification F1 (Arg-C), where a predicted event argument is considered correct only if both its boundaries and role type are accurate. We conduct experiments on 5 runs with different seeds and report the average results.



\noindent \textbf{Baselines} \quad
Our baselines include: (1) Two SOTA span-based methods, \textit{TSAR}~\cite{xu2022two} and \textit{SCPRG}~\cite{liu2023enhancing};
(2) Five typical generation-based methods, \textit{DocMRC}~\cite{liu-etal-2021-machine}, \textit{EEQA}~\cite{du-cardie-2020-event}, \textit{FEAE}~\cite{weietal2021trigger}, \textit{BART-Gen}~\cite{lietal2021document}, and \textit{HRA}~\cite{ren2023retrieve};
(3) Four prompt-based approaches: \textit{RKDE}~\cite{hu2023role}, \textit{PAIE}, SPEAE~\cite{nguyen2023contextualized} and \textit{TableEAE}.  We also compare with \textit{LLM approaches} and conduct a detailed analysis in the Appendix~\ref{Sec:Comparison with Large Language Models}.

Most baselines are originally proposed as Single-EAE methods. For the Multi-EAE baselines, we extend PAIE and TabEAE to obtain PAIE-multi\footnote{We extend the original PAIE~\cite{ma2022prompt} into the multi-EAE framework by annotating triggers within the context and concatenating prompts for multiple events. The rest of the approach remains consistent with the original PAIE.} and TabEAE-multi\footnote{TabEAE~\cite{he2023revisiting} aims to capture event co-occurrence and trains the model on multi-event scheme but infers on single-event scheme. For a fair comparison, we train their method on multi-event instances and conduct inference on multi-event instances as well.}.  Our experimental details are shown in Appendix~\ref{Sec:Experimental Details}.


\subsection{Main Results}
\label{main results}

Table~\ref{tab:main_results} illustrates the performance comparison between our proposed DEEIA method and various baseline approaches across three datasets. {(Experimental results on ACE05 dataset are shown in Appendix~\ref{sec:A3}.)} Our approach consistently achieves optimal results across all datasets generally, irrespective of the evaluation metric employed.
Further analyzing the experimental result data, we observe: (1) 
The performance of DEEIA outperforms that of Single-EAE baselines. This indicates our DEEIA can \textbf{effectively capture the beneficial event correlations}, enhancing the performance of EAE task. 
 (2) 
 PAIE-multi exhibits significantly lower performance compared to PAIE, which illustrates that  simultaneously processing multiple events significantly increases the difficulty of the task.  While our DEEIA significantly outperforms the baseline  PAIE-multi on three datasets, demonstrating that our DEEIA effectively addresses the challenge of multi-event information complexity.  
 (3) {Compared to PAIE and TabEAE, the improvement of our DEEIA model on RAMS is around 0.9-1.2 F1, while on WikiEvents and MLEE, the improvement is around 1.2-1.7 F1 and 1.7-3.7 F1 respectively.} Therefore, the improvement of DEEIA on WikiEvents and MLEE is more pronounced compared to RAMS. We hypothesize this may be because  WikiEvents and MLEE contain a higher proportion of multi-event instances, in contrast to the RAMS dataset, which is primarily composed of single-event instances. (Distributions of event numbers on three datasets are shown in Appendix~\ref{appendix:data_statictics}.)



\begin{table*}[htbp]
\centering
\small{
\setlength{\tabcolsep}{2.5mm}{
\centering
\begin{tabular}{lcccccc}
\hline
\multirow{2}{*}{Model}  & \multicolumn{3}{c}{RAMS} & \multicolumn{3}{c}{WikiEvents}   \\  \cline{2-7} 
& \small{All [871]}  & \small{\# E = 1 [587]}    & \small{\# E > 1 [284]}   & \small{All [365]} & \small{\# E = 1 [114]}      & \small{\# E > 1 [251]}  \\
\hline                
PAIE-multi          & {50.86±0.22}        & {51.75±0.34}        & {48.94±0.33}          & 61.74±0.62 & 65.01±1.20 & 60.18±0.81                \\
TabEAE-multi                   
              & 51.44±0.32        & 52.27±0.28       & 50.42±0.44               &   65.68±0.62 & 67.08±0.44 & 65.02±0.28       
                        \\
DEEIA(Ours)         & \textbf{53.36±0.44}         & 53.64±0.60        & \textbf{52.76±0.32}          & \textbf{66.95±0.66}     & \textbf{67.49±0.72} & \textbf{66.57±0.62}    \\  
\quad \small{w/o DE}         &  52.38±0.49
        & \textbf{53.84±0.74}
        &    49.86±0.78
       & {64.90±1.07}     & {66.96±1.06} & 63.79±0.77    \\  
\quad \quad \small{w/o intra}         &  51.49±0.55
        & 53.06±0.62
        &    48.34±0.64
       & {64.65±0.56}     & {66.86±0.44} & 63.12±0.75    \\ 
\quad \quad \small{w/o inter}         &  52.18±0.44
        & 53.56±0.52
        &    49.16±0.58
       & {65.65±0.66}     & {67.16±0.48} & 65.00±0.66    \\  
\quad \small{w/o PE}         &  52.41±0.37
       &  53.08±0.43
       &  51.06±0.38
         &  {66.02±0.70}    & 67.06±0.58 & 65.59±0.72    \\  
\quad \small{w/o EIA}        & 51.70±0.56
         & 52.31±0.52
        &  49.90±0.68
        & {64.20±0.24}     & 65.43±0.42 & 63.57±0.20    \\  
\quad \small{w/o DE \& EIA}         &  51.25±0.58
       &  52.14±0.54
      &  49.66±0.62
         & {62.37±1.04}     & 65.56±1.06 & 61.68±0.88    \\  \hline
\end{tabular}
}
\caption{Ablation study on RAMS and WikiEvents. Strict argument classification F1 scores (Arg-C) are reported. \# E means the number of events in an instance and $\text{[]}$ indicates the number of instances of this kind.  \textbf{Bold} indicates the best experimental results.  The reported results are averaged from 5 different random seeds.}
\label{tab:ablation study}
}
\end{table*}

\subsection{Ablation Study}
To better illustrate the effectiveness of different components, we conduct ablation studies on RAMS and WikiEvents datasets in Table~\ref{tab:ablation study}. We divide the dataset into two parts based on the number of events in each instance: those with \# E > 1 and those with \# E = 1, and report the Arg-C F1 scores to explore the impact of our modules on instances with single and multiple events. 

\textbf{Without Dependency-guided Encoding (DE)}. We replace the Dependency-guided Encoding (DE) module with a vanilla transformer encoder. 
This results in performance reduction, primarily attributed to the decline in the performance of  multi-event samples, which illustrates that DE module effectively provides dependency guidance for multi-event extraction.
We further explore the effectiveness of intra and inter dependencies.  It is observed that both intra and inter dependencies contribute positively to the model. When remaining only one type of dependency, the intra dependency has a beneficial effect on the model, but the inter dependency has a negative effect.

\textbf{Without Event-specific Information Aggregation (EIA)}.
The performance of both multi-event and single-event samples has significantly declined on two datasets. This indicates that our EIA module can provide beneficial event-specific information. Moreover, when removing both DE and EIA modules, the performance decay exceeds that when removing a single module, which explains that our two modules can work together. 

\textbf{Without Prompt Enhancing (PE)}.
When removing the event type information defined in \S~\ref{Prompt Enhancement.}, the performance on two datasets decays slightly, which indicates that event type helps the model distinguish between different event prompts and integrates wider information.


\section{Analysis}

\subsection{Efficiency Analysis}
Table~\ref{complexity} reports the efficiency of different prompt-based methods. First, compared to Single-EAE baseline PAIE, our method saves 8.76\%, 32.96\%, and \textbf{35.20\% } inference time  on RAMS, WikiEvents, MLEE datasets respectively.
This fully demonstrates the efficiency superiority of our Multi-EAE approach. Additionally, our method significantly reduces 
55.57\%, 57.38\% and \textbf{69.19\%} inference time compared to TabEAE-multi, with almost no increase in the number of parameters. This illustrates that our DEEIA model enhances the efficiency of the document-level EAE task.

\begin{table}[tbp]
\centering
\small{
\setlength{\tabcolsep}{0.7mm}{
\begin{tabular}{lcccc}
\hline
\centering
\multirow{2}{*}{Method} & \multirow{2}{*}{Params} & \multicolumn{3}{c}{Inference Time} \\ \cline{3-5} 
                        &                         & \small{RAMS}     & \small{Wikievents}    & \small{MLEE}    \\ \hline
PAIE            & 406.21M                  & 15.86    & 8.83         & 14.29   \\
PAIE-multi             & 406.21M                  & 12.53    & 5.15         & 8.57   \\
TabEAE-multi            & 383.78M                  & 32.59    & 13.89         & 30.06   \\
DEEIA (Ours)              & 388.12M                  & 14.47    & 5.92          & 9.26   \\
 \hline
\end{tabular}
}
\caption{Inference time (second) for different models (large) on
test set of three datasets. Experiments
are run on one same Tesla A100 GPU.}
\label{complexity}
}
\end{table}


\subsection{Effect 
Analysis on Event Numbers}
\begin{figure}[tbp]
    \centering
    \includegraphics[width=1.0\linewidth]{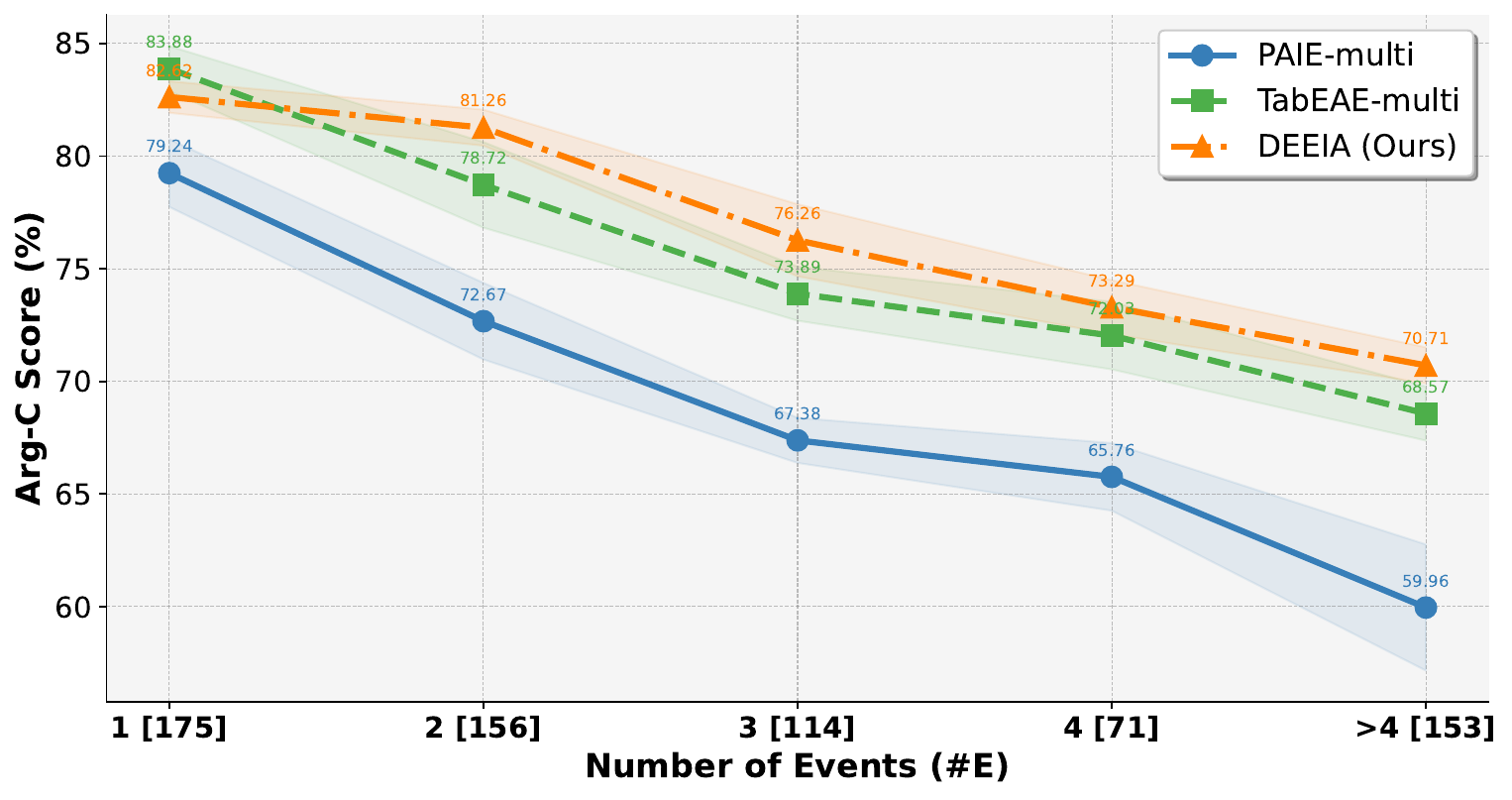}
    
    \caption{The averaged performance of the PAIE-multi, TabEAE-multi, and DEEIA models on samples with different event numbers in  MLEE dataset. Our model achieves better results on samples with multiple events.}
    \label{fig:mlee}
\end{figure}

To further investigate the effectiveness of our method in addressing the multi-event information complexity problem, we divide the documents in the development sets of WikiEvents and MLEE into different groups based on the event numbers\footnote{We do not use the RAMS dataset because the RAMS dataset has  a low proportion of multi-event instances.}. As illustrated in Figure~\ref{fig:mlee}, as the event number increases, we observe a decreasing trend in the performance of all models. We believe this is due to the fact that more events require the model to process more complex information and longer text, which is more difficult. Furthermore, we find that the baseline model PAIE-multi performs significantly worse on samples where the number of events exceeds two. In contrast, our model demonstrates a marked improvement in multi-event samples compared to PAIE-multi and TabEAE-multi, which shows the superiority of DEEIA in capturing the event correlations among multiple events. The results on WikiEvents dataset are in Appendix~\ref{Sec:event_nums_wiki}.

\subsection{Analysis of Two Modules}
 \textbf{Dependency Guidance} \quad To investigate how the attentive biases influence the self-attention  mechanism, we visualize all attentive biases (calculated in Eq.~\ref{eq4}) for the test sets of all three datasets.  We conducted a detailed analysis in Appendix~\ref{Sec:Analysis of Dependency Guidance}. We find that both inter and intra dependencies can provide effective information guidance for Multi-EAE task. In datasets with a larger proportion of multi-event documents, such as the MLEE dataset, both inter and intra dependencies exhibit significant effects, while in datasets with a smaller proportion of multi-event documents, such as RAMS, intra dependency plays a primary role.

\begin{figure}[tbp]
    \centering
    \includegraphics[width=1.0\linewidth]{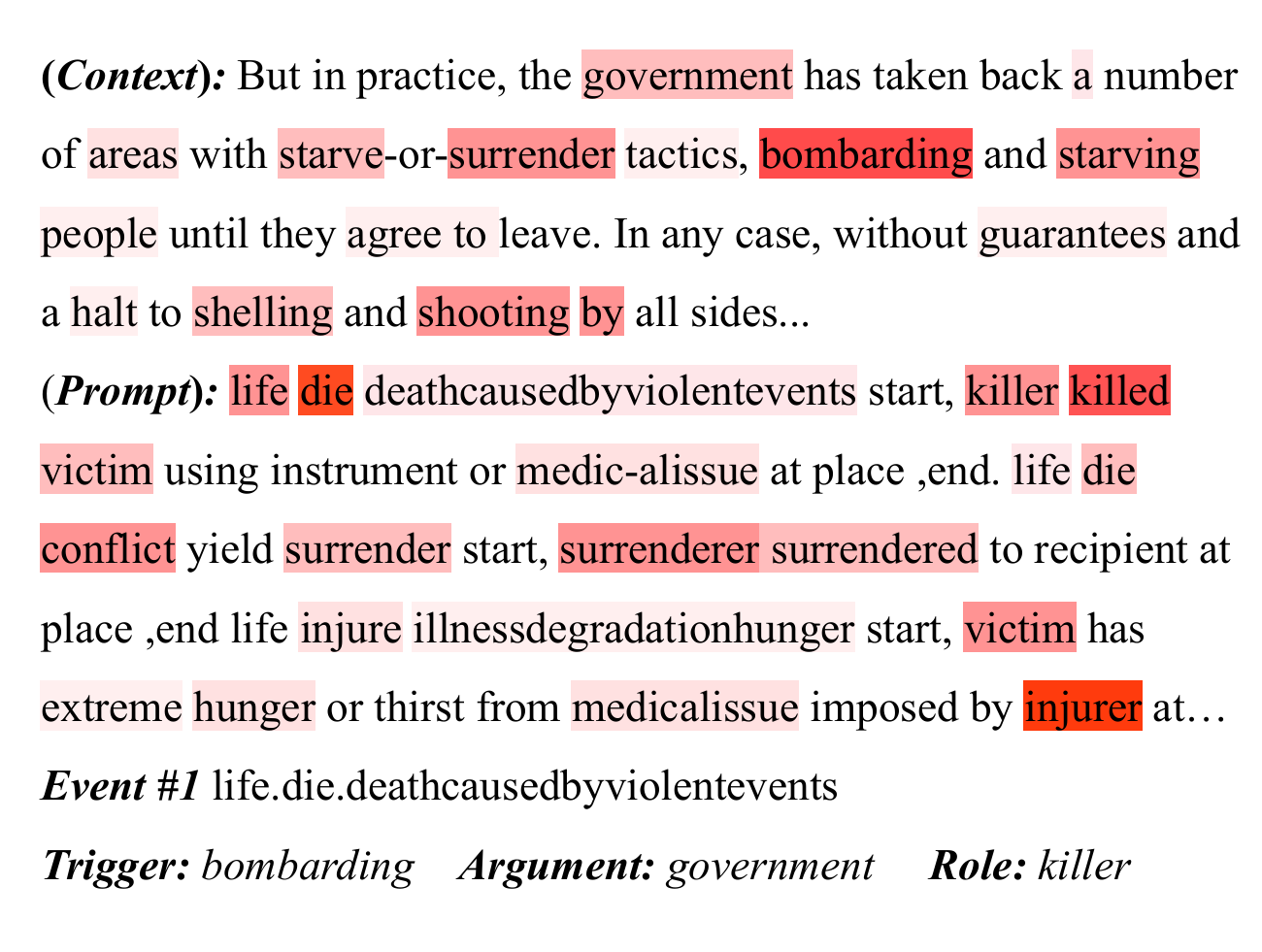}
    
    \caption{Visualization of attentive weights in EIA module from an example in RAMS. We calculate the attentive weight $\textbf{p}_{k}$ based on the representations of argument slot ``government'' and the trigger ``bombarding''.  }
    \label{fig:fig_pooling}
\end{figure}

\textbf{Analysis of Information Aggregation} \quad
To assess the effectiveness of our EIA module in capturing event-specific contextual information, we visualize the attentive weights $\textbf{p}_{k}$ in Eq.~\ref{eq6} of an argument ``government'' of Figure~\ref{fig:fig1}. As shown in Figure~\ref{fig:fig_pooling}, our
EIA module gives high weights to the context words, such as
\textit{starving}, \textit{shooting} and \textit{surrender}, prompt words such as \textit{die}, \textit{killer} and \textit{injurer}, which benefits the argument extraction of ``government''.  
Interestingly, some words such as \textit{surrender} and \textit{shelling} also act as triggers or arguments in other events, which  reveals the EIA module's capability to capture event correlations. 

\subsection{Error Analysis and Case Study}

\textbf{Error Analysis} \quad We  further conduct error analysis to explore the effectiveness of our DEEIA model in 
Appendix~\ref{Sec:error_analysis}. We analyze all prediction errors on WikiEvents test set and categorize them into five classes. As shown in Figure~\ref{fig:error_analysis}, compared to the Single-EAE baseline PAIE, our DEEIA model reduces the number of errors from 312 to 292, indicating the effectiveness of DEEIA in capturing event correlations. Compared to PAIE-multi, DEEIA reduces the number of errors from 358 to 292. Additionally,  our DE and EIA modules also significantly reduce specific types of errors. The detailed analysis is shown in 
Appendix~\ref{Sec:error_analysis}

\begin{figure}[tbp]
    \centering
    \includegraphics[width=1.0 \linewidth]{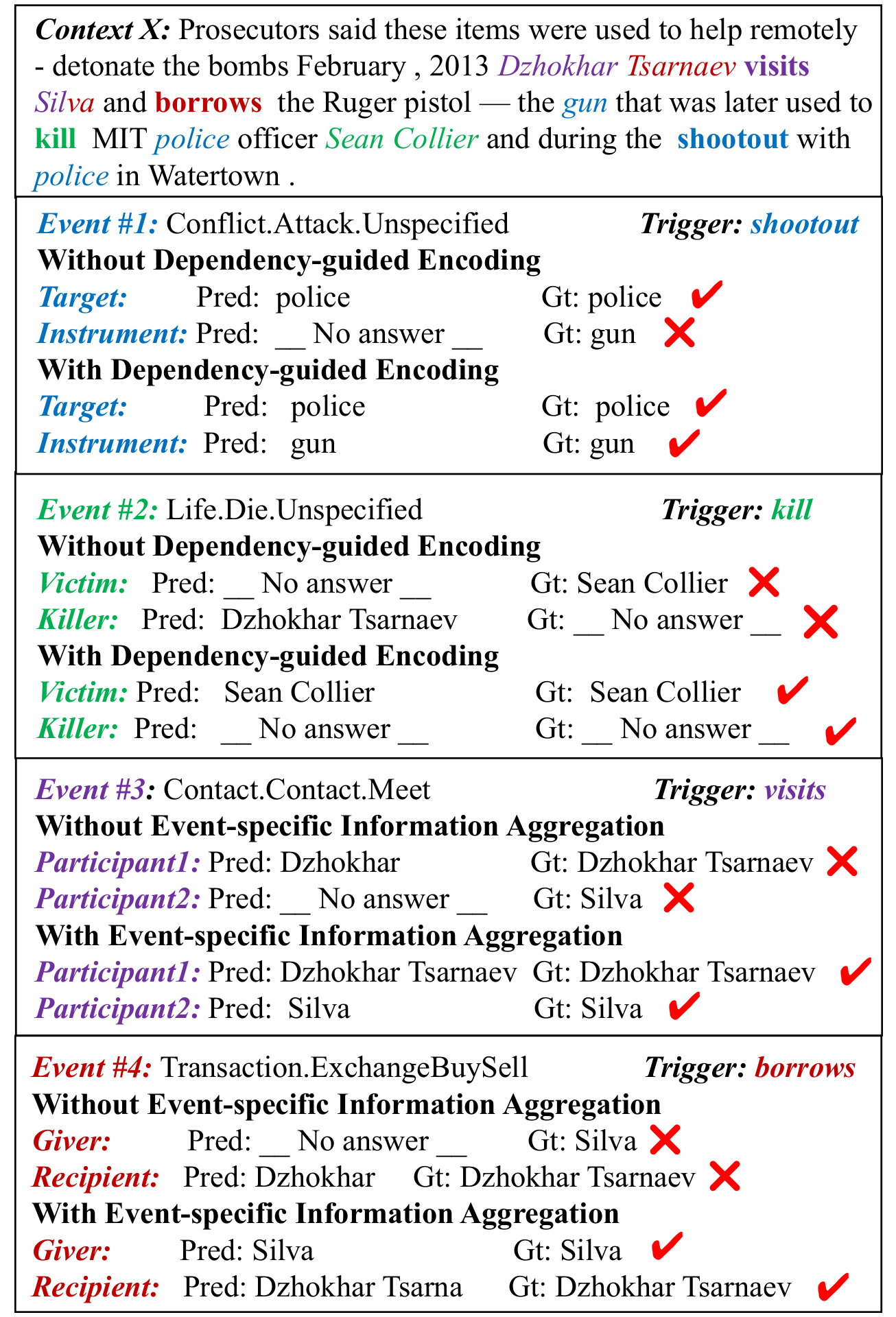}
    
    \caption{A multi-event test case from WikiEvents.}
    \label{fig:case_study}
\end{figure}

\textbf{Case Study} \quad We conducted the case study to further explore the effect of our proposed modules in multi-EAE. As shown in Figure~\ref{fig:case_study}, this is a complex document containing four events and there exists the argument overlapping phenomenon. First, without the  dependency-guide encoding (DE), our model fails to identify arguments such as  ``Sean Collier'' and ``gun''. However, with the DE module, our model correctly predicts the roles of these arguments. This demonstrates that the event dependencies provide beneficial guidance. Additionally, with the EIA module, our model is capable of extracting overlapping arguments like ``Dzhokhar Tsarnaev'' and ``Silva'', which indicates that the EIA module provides event-specific contextual information for a
better context understanding.

\section{Conclusion}
In this paper, we propose a Multi-EAE  model DEEIA, which overcomes the inefficiency limitations of traditional EAE methods. The proposed Dependency-guide Encoding (DE) module and Event-specific Information Extraction (EIA) module effectively enhances the model's ability to understand complex multi-event contexts. Our extensive experiments on three public benchmarks illustrate the superiority of our model in performance and efficiency. 
\label{sec:bibtex}

\section{Limitations}
The primary limitation of our method is the issue of input length. Concatenated text and prompts are more easy to exceed the maximum input length. Currently, our solution to this is to use a sliding window approach~\cite{zhou2021document, zhang2021document} to encode the sequences of different windows and average the overlapping token embeddings of different windows to obtain the final representation. However, this method is not the optimal solution for processing long texts, which leads to the information loss and results in suboptimal performance. Therefore, in the future, we will explore to address the challenge of long input text with the aim of enhancing our DEEIA model.

\section*{Acknowledgements}
This work was supported by the National Science Foundation of China under Grant 62106038, and in part by the Sichuan Science and Technology Program under Grant 2023YFG0259. 
\bibliography{anthology,custom}
\bibliographystyle{acl_natbib}

\clearpage

\appendix

\begin{table}[ht]
\centering
\setlength{\tabcolsep}{1.0mm}{
\begin{tabular}{@{}lccc@{}}
\hline
\textbf{Dataset}      & \textbf{RAMS} & \textbf{WikiEvents} & \textbf{MLEE} \\ \hline
\textbf{\# Event types}        & 139  & 50         & 23   \\
\textbf{\# Events per text}   & 1.25 & 1.78       & 3.32 \\
\textbf{\# Args per event}  & 2.33 & 1.40       & 1.29 \\
\midrule
\textbf{\# Events}              &      &            &      \\
Train               & 7329 & 3241       & 4442 \\
Dev                  & 924  & 345        & -    \\
Test                 & 871  & 365        & 2200 \\
\bottomrule
\end{tabular}
}

\caption{Dataset Statistics.}
\label{tab:dataset Statistics}
\end{table}

\begin{algorithm}
\caption{Multi-EAE with Dynamic Windows}
\begin{algorithmic}[1]
\Require Input context $X$, concatenated multi-event prompts $P$, window sizes $d_1$ and $d_2$ ($d_1 + d_2 < \text{max length}$)
\Ensure Final representation of the sequence
\If{$\text{length}(X + P) > \text{max length}$}
    \State  Split $X$ into $\{X_1, X_2, \ldots, X_n\}$ using $d_1$
\EndIf
\For{each $X_i$ in $\{X_1, X_2, \ldots, X_n\}$}
    \State  Identify number of events in $X_i$ based on triggers, get $P_i$
    \If{$\text{length}(X_i + P_i) > \text{max length}$}
        \State  Split $P_i$ into $\{P^1_i, \ldots, P^m_i\}$ using  $d_2$
    \EndIf
    \For{each $P^j_i$ in $\{P^1_i, \ldots, P^m_i\}$}
        \State Concatenate $X_i$ with $P^j_i$
        
        \State Encode them to $S^j_i$
    \EndFor
    \State Average pool $S^j_i$ to obtain final $S_i$
\EndFor
\State \textbf{Aggregate:} Pool $S_i$ to get the final sequence representation
\end{algorithmic}
\label{alg1}
\end{algorithm}

\section{Dynamic Window Algorithm}
{
For sequences that surpass the maximum length of 512, we employ a sliding window approach to process longer sequences. For the general case of processing long input texts, we have designed the following Algorithm~\ref{alg1} based on sliding windows. 
 }
{In our implement, PAIE~\cite{ma2022prompt} has already utilized a sliding window to divide the long document into several instances. We only use the sliding window 
 to process the prompts and ultimately obtain the final sequence representation through pooling. In our experiments, both $d_1$
 and $d_2$
 are set to 250.}

\section{Experimental Details}
\subsection{Dataset Statistics}
\label{appendix:data_statictics}
We evaluate our proposed method on four  event argument extraction datasets.

\textbf{RAMS}~\cite{ebner2020multi} is a document-level EAE dataset with 9,124 annotated events from English online news, annotated event-wise. Following ~\cite{he2023revisiting}, we employ a sliding window approach to aggregate events in the same context into single instances with multiple events, following the original train/dev/test split.
 
\textbf{WikiEvents}~\cite{zhang2020two} is a document-level EAE dataset featuring events from English Wikipedia and associated news articles. It includes co-reference links for arguments, but we only use the exact argument annotations in our experiments.

\textbf{MLEE}~\cite{10.1093/bioinformatics/bts407} is a document-level event extraction dataset, contains manually annotated abstracts from bio-medical publications in English. We follow the preprocessing steps outlined by~\cite{trieu2020deepeventmine}. Since there is only train/test
data split for the preprocessed dataset, we employ
the training set as the development set.

{\textbf{ACE05}~\cite{doddington2004automatic} is a labeled corpus used for information extraction, consisting of newswire, broadcast news, and telephone conversations. We employ its English event annotations for sentence-level Event Argument Extraction (EAE). The data preprocessing follows the method described by ~\cite{ma2022prompt}.}

\begin{figure*}[tbp]
    \centering
    \subfloat[\label{fig:data_rams}RAMS]{
    \includegraphics[width=0.32\linewidth]{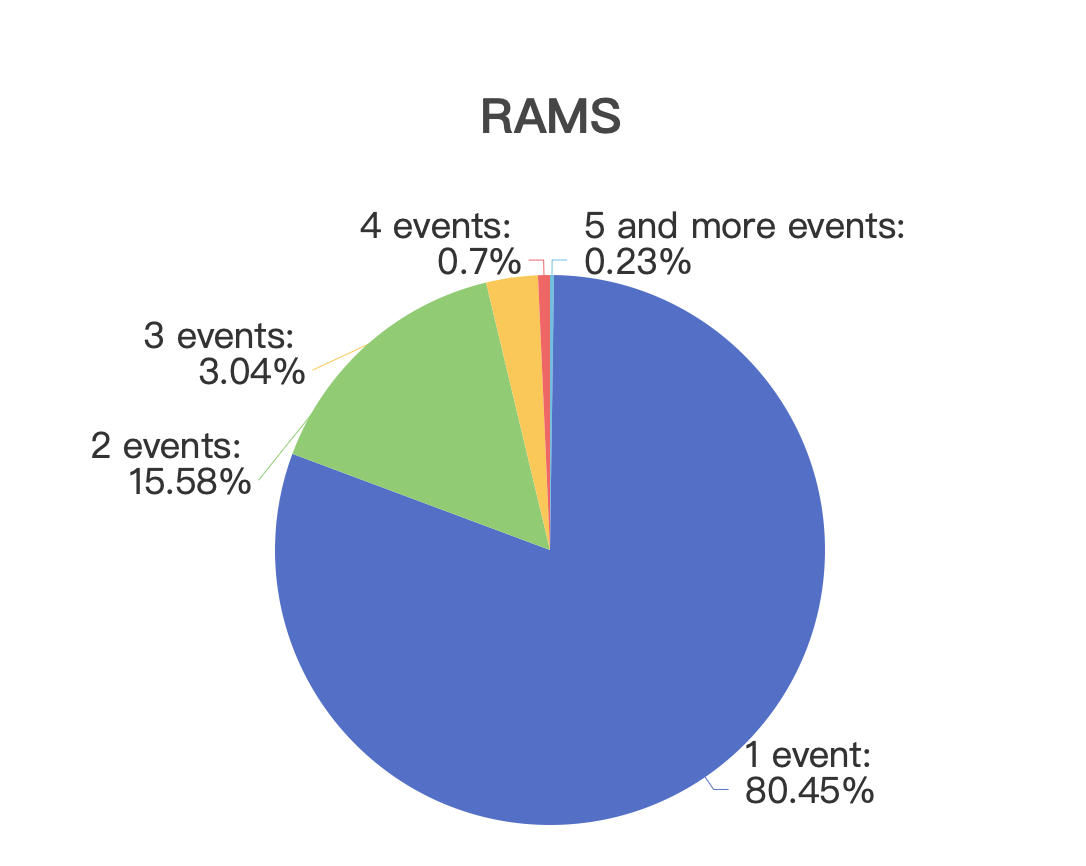}
    }
    \subfloat[\label{fig:data_wiki}WikiEvents]{
    \includegraphics[width=0.32\linewidth]{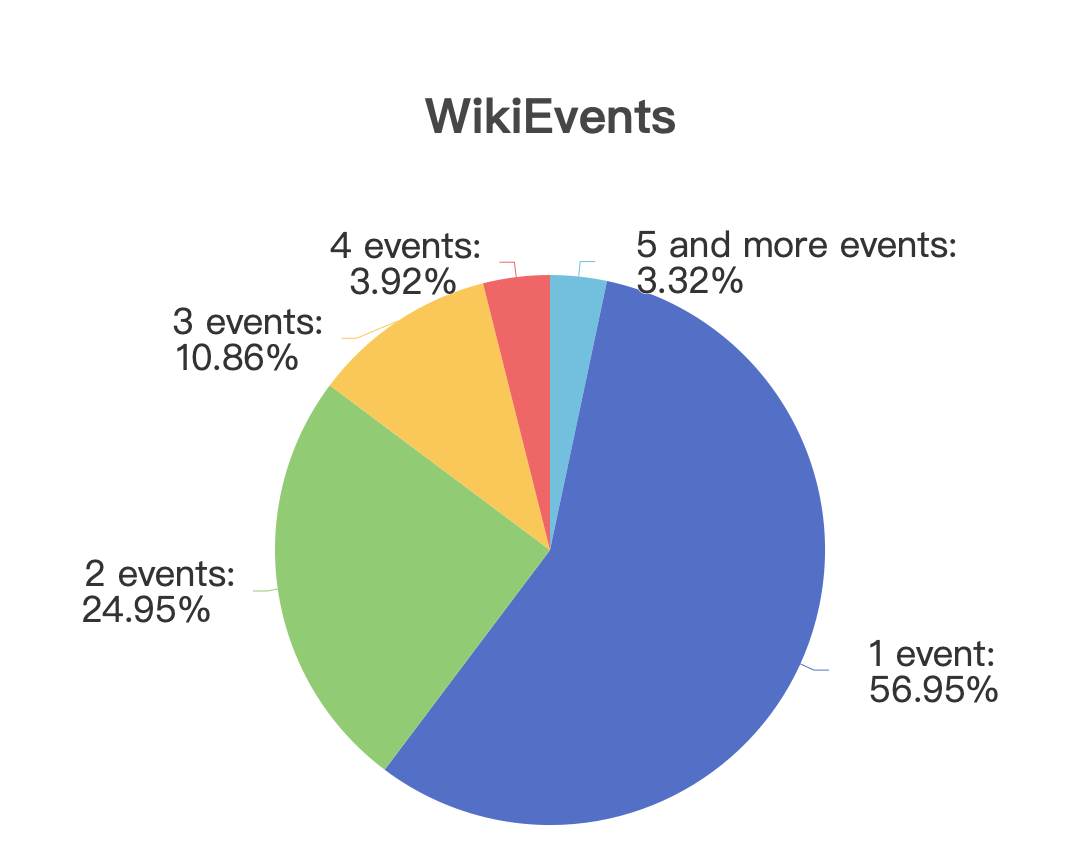}
    }
    \subfloat[\label{fig:data_mlee}MLEE]{
    \includegraphics[width=0.32\linewidth]{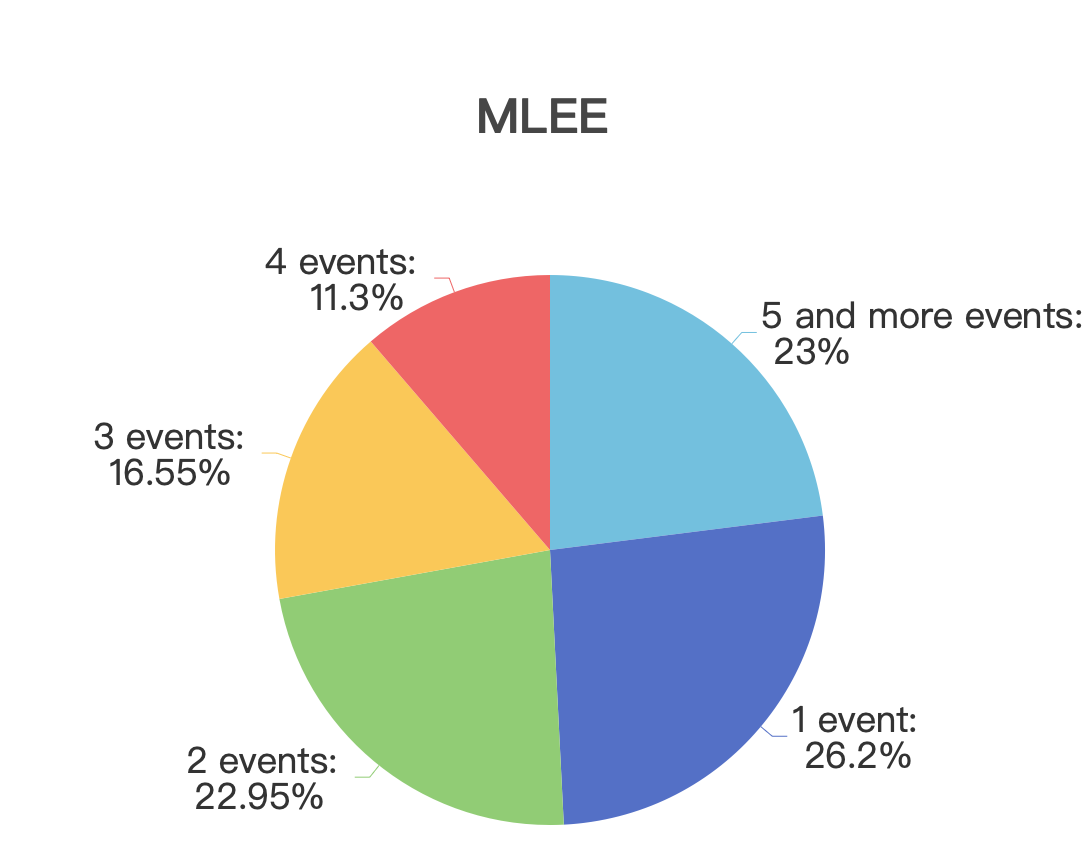}
    }

    \caption{Distributions  of the number of events per
instance on the three document-level datasets. }
\label{fig:data_percent}
    
\end{figure*}

The detailed dataset statistics of three datasets are shown in Table~\ref{tab:dataset Statistics}. We also calculate the  distributions of the number of events per instance on the three dataset, which are shown in Figure~\ref{fig:data_percent}.  As shown in Figure~\ref{fig:data_percent},  three datasets exhibit different data distributions between single-event samples and multi-event samples. For RAMS dataset, single events samples dominate the majority, while the proportion of multi-event samples is quite small. However, for WikiEvents and MLEE datasets, multi-event samples account for a significant proportion.
\subsection{Experimental Details}
\label{Sec:Experimental Details}
{According to TabEAE, using RoBERTa as the PLM outperforms BART across multiple approaches (such as PAIE and TabEAE). Therefore, we adopt RoBERTa as our PLM so as to compare to the SOTA method.}
Our implementation utilizes Pytorch and runs on a Tesla A100 GPU. We configure the encoder using the initial 17 layers of RoBERTa-large~\cite{liu2019roberta}. The decoder's self-attention and feedforward layers inherit their weights from RoBERTa-large's subsequent 7 layers. This division of a 17-layer encoder and a 7-layer decoder is empirically determined as the most effective configuration~\cite{he2023revisiting}. It's important to note that the decoder's cross-attention component is initialized randomly, with its learning rate set at 1.5 times that of other parameters.  More detailed hyper-parameter setting is shown in Table~\ref{table:hyper}. We utilize the prompts proposed in PAIE~\cite{ma2022prompt}, which are shown in Table~\ref{tab:prompt}.

\begin{table}[h]
\centering
\setlength{\tabcolsep}{0.3mm}{
\begin{tabular}{lccc}
\hline
\textbf{Hyperparameters}  & \textbf{RAMS} & \textbf{Wiki} & \textbf{MLEE} \\ \hline
Training Steps                                          & 10000         & 10000              & 10000         \\
Warmup Ratio*                                               & 0.1           & 0.1                & 0.2           \\
Learning Rate*                 & 2e-5          & 3e-5               & 3e-5          \\
Max Gradient Norm                & 5             & 5                  & 5             \\
Batch Size*                         & 4             & 4                  & 4             \\
Context Window Size              & 250           & 250                & 250           \\
Max Span Length                  & 10            & 10                 & 10            \\
Max Encoder Seq Length           & 500           & 500                & 500           \\
Max Prompt Length*                & 210          & 360                & 360          \\
Encoder Layers*                & 17          & 17                & 17 \\
Decoder Layers*                & 7          & 7                & 7 \\
Gamma*              & 0.01          & 0.1                & 0.1
\\ \hline
\end{tabular}}
\caption{Hyperparameter settings. * means that we tuned the hyperparameters in our experiments. The rest of hyperparameters are set the same as PAIE~\cite{ma2022prompt}. }
\label{table:hyper}
\end{table}

\subsection{Experimental Results on ACE05}
\label{sec:A3}
{
We evaluate our proposed model on the ACE05 dataset~\cite{doddington2004automatic}, and the specific experimental results are shown in Table~\ref{tab:main_results_new} (The reported results are averaged from 5 different random seeds). Experimental results demonstrate that our proposed DEEIA model also performs well on the sentence-level ACE05 dataset, significantly improving the extraction performance of multi-event instances.
}
\begin{table}[htbp]
\centering

\small{
\setlength{\tabcolsep}{2.5mm}{
\begin{tabular}{llcc}
\hline
\multirow{2}{*}{Method}  
& \multirow{2}{*}{PLM}
& \multicolumn{2}{c}{ACE05} \\ \cline{3-4} 
                              &    & Arg-I & Arg-C \\ \hline

${\text{EEQA}}$~(\citeyear{du-cardie-2020-event})   & ${\text{BERT-l}}$    & 70.5 & 68.9      \\ 
 
${\text{EEQA}}$~(\citeyear{du-cardie-2020-event})   & ${\text{RoBERTa-l}}$    & 72.1  & 70.4      \\

 ${\text{BART-Gen}}$~(\citeyear{lietal2021document}) & ${\text{BART-l}}$ & 69.9     & 66.7          \\

${\text{PAIE}}$~(\citeyear{ma2022prompt})   & ${\text{BART-l}}$ & 75.7     & 72.7        \\ 
${\text{PAIE}}$~(\citeyear{ma2022prompt})   & ${\text{RoBERTa-l}}$ & 76.1     & 73.0        \\ 

${\text{TabEAE}}$~(\citeyear{he2023revisiting})  & ${\text{RoBERTa-l}}$ & 75.9     & 73.4         \\

\hline
$\text{DEEIA (Ours)}$    & ${\text{RoBERTa-l}}$
&\textbf{76.3} & \textbf{74.1}             \\ \hline
\end{tabular}
}
\caption{
{
Comparison of performance on ACE05 test set. * means we rerun their code based on their experimental settings. 
\textbf{Bold}  indicates the best  experimental results. 
}}

\label{tab:main_results_new}
}
\end{table}

\section{Experimental Analysis}
\subsection{Effect 
Analysis on Event Numbers in WikiEvents}
\label{Sec:event_nums_wiki}
\begin{figure}[tbp]
    \centering
    \includegraphics[width=1.0\linewidth]{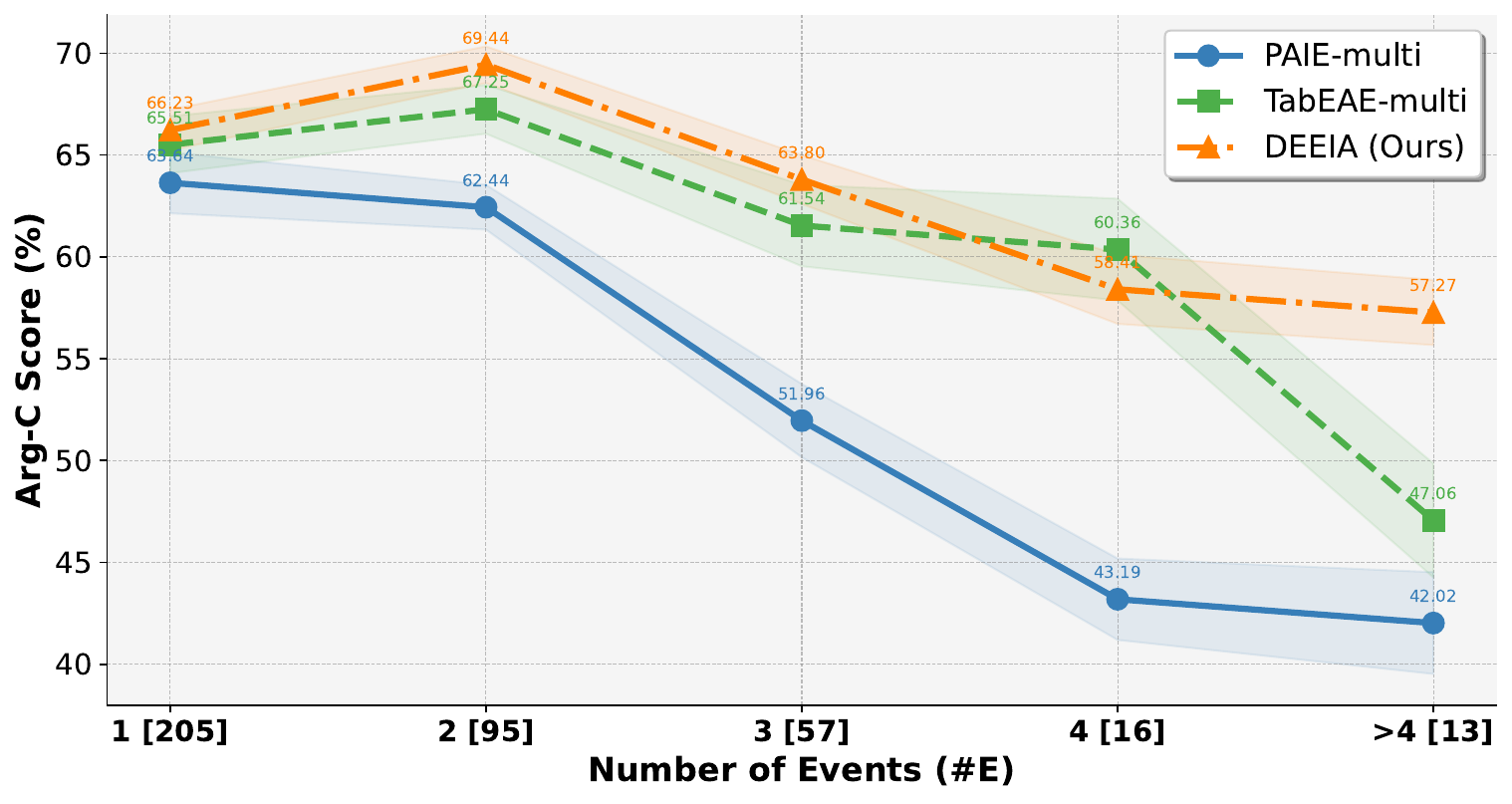}
    
    \caption{The averaged performance of the PAIE, TabEAE, and DEEIA models on samples with different event numbers in the WikiEvents dataset. Our model achieves better results on samples with multiple events. }
    \label{fig:wiki}
\end{figure}

As illustrated in Figure~\ref{fig:wiki}, as the event number increases, we observe a decreasing trend in the performance of all models. We believe this is due to the fact that more events require the model to process more complex information and longer text, which is more difficult. Furthermore, we find that the baseline model PAIE performs significantly worse on samples where the number of events exceeds two. In contrast, our model demonstrates a marked improvement in multi-event samples compared to PAIE-multi and TabEAE-multi, which shows the advantages of DEEIA in Multi-EAE.

\subsection{Analysis of Dependency Guidance}
\label{Sec:Analysis of Dependency Guidance}
\begin{figure}[tbp]
    \centering
    \includegraphics[width=1.0\linewidth]{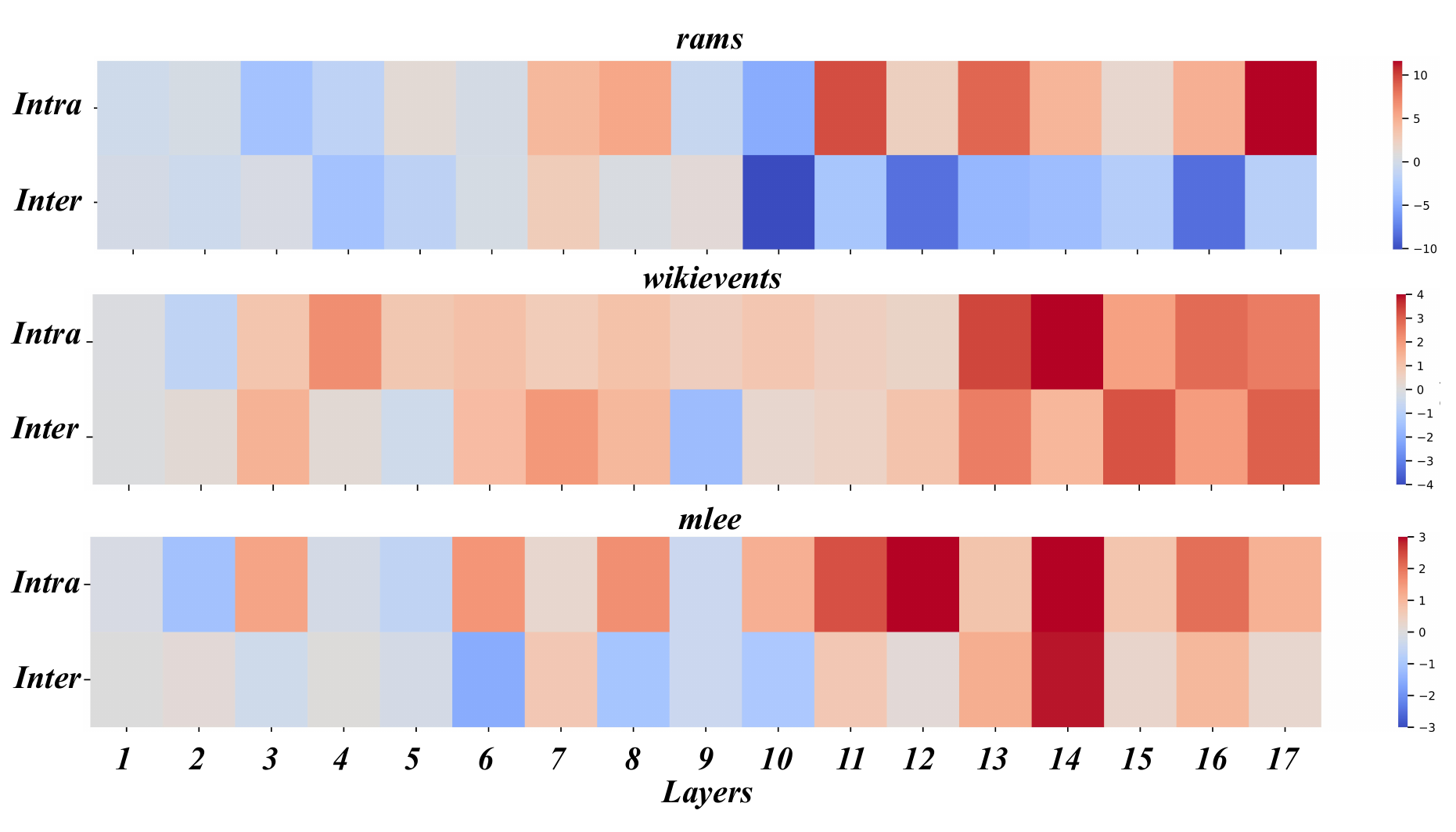}
    
    \caption{Visualization of the learnable attentive biases, where each cell represents the value of an attention bias. The horizontal axis represents intra-event and inter-event dependencies, and the vertical axis indicates each transformer layer incorporating structural guidance. }
    \label{fig:fig_structural_guidance}
\end{figure}

To investigate the manner in which the learnable attentive biases influence the self-attention  mechanism, we collect all attentive biases (calculated in Eq.~\ref{eq4}) for the test sets of all three datasets. These biases are then categorized based on dependency types and averaged across all attention heads and instances. As shown in Figure~\ref{fig:fig_structural_guidance}, 
the self-attention scores are primarily determined by vanilla self-attention, with minimal influence from dependency information at bottom layers. However, as the number of layers increases, the impact of learnable attentive biases gradually becomes significant, especially between layers 12 to 16.

Additionally,  we observe that for different datasets, the attentive bias distributions corresponding to inter-event and intra-event dependencies are different. For  RAMS dataset, the attentive bias associated with intra-event dependency is relatively positive, but that corresponding to inter-event dependency is relatively negative. We believe that in the RAMS dataset, there are more single-event samples, and the model focuses more on learning intra-event information associations. However, for WikiEvents and MLEE datasets, the attentive biases for both dependencies are mostly positive, indicating that in these datasets, samples with multiple events are more prevalent, and both dependencies provide beneficial guidance for the model to solve the information complexity problem.
\subsection{Architecture Variants}
\label{sec: Architecture Variants}
In this section, we explore the effect of different architectures and define two types of architecture variants.
(1) To explore whether to integrate the dependency information into the decoder, we define $\textbf{DEEIA}_B$, which integrates the dependency information to both the encoder and decoder. (2)
Since the prompts for events are defined based on their event types, the same event types will have the same prompts. Therefore, in a document, if there are multiple events of the same type, whether to concatenate repeated prompts becomes an option. We concatenate repeated prompts, and call this $\textbf{DEEIA}_M$. 

As shown in Table~\ref{Analysis of variant architectures.}, there is a minor performance drop across all three datasets when dependency information is integrated into the decoder. This implies that embedding dependency information during the encoding phase is adequate, and overloading the model with excessive informational guidance is not of benefit.
Furthermore, concatenating repeated prompts results in a slight improvement in the performance for the RAMS and WikiEvents datasets, but a marginal decline for the MLEE dataset. Overall, the improvement is not substantial. We believe that while concatenating repeated prompts increases prompt diversity, it also extends the sequence length, thereby increasing the difficulty of long-distance reasoning.

\begin{table}[ht]
\centering

\setlength{\tabcolsep}{2.7mm}{
\begin{tabular}{lccc}
\hline
Method      & RAMS & WikiEvents & MLEE \\ \hline
$\text{DEEIA}$       & 53.4  & 67.0         & 74.3   \\
$\text{DEEIA}_M$   & 53.5 & 67.2       & 74.4 \\
$\text{DEEIA}_B$  & 53.2 & 66.5       & 74.6 \\
\hline
\end{tabular}
}
\caption{Analysis of variant architectures.}
\label{Analysis of variant architectures.}
\end{table}

\begin{figure*}[tbp]
    \centering
    \includegraphics[width=1.0\linewidth]{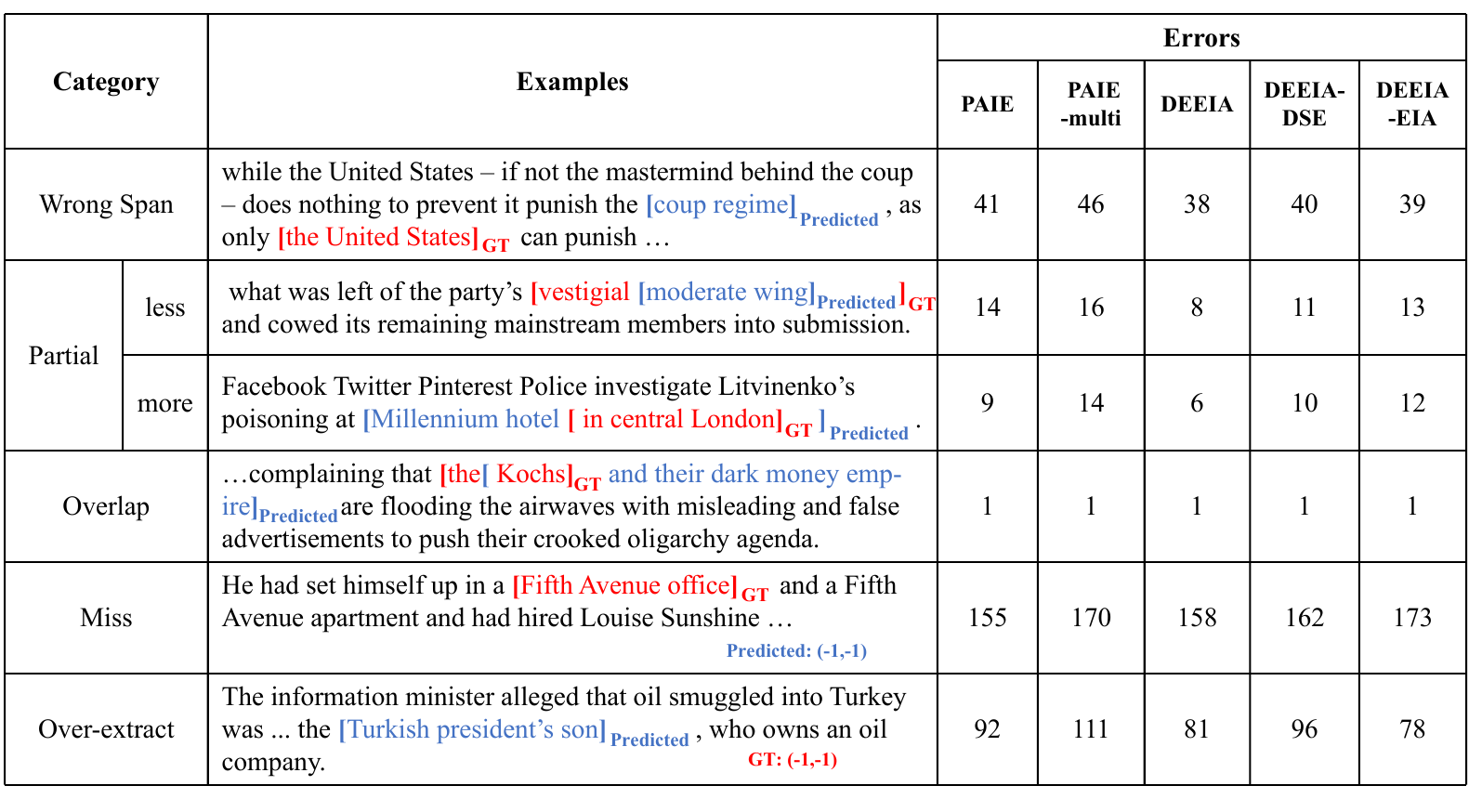}
    
    \caption{Error Analysis on WikiEvents test set. We summarize the errors into five categories and count the number of errors for different models. Blue represents the model's predictions, while red represents the ground truth.}
    \label{fig:error_analysis}
\end{figure*}



\subsection{Error Analysis}
\label{Sec:error_analysis}
To compare different models in greater detail, we conduct error analysis on dev sets of RAMS, WikiEvents and MLEE datasets. We divide the errors into five categories,
which is shown in Figure~\ref{fig:error_analysis}. \textbf{Wrong span} refers to the case where the predicted span and the true span have no intersection; \textbf{Partial} refers to the case where the predicted span and the golden span partially overlap, which means the predicted span is a proper subset of the golden span or vice versa; \textbf{Overlap} occurs when there is a non-partial case, indicating that there is overlap between the predicted span and the golden span; \textbf{Over-extraction} refers to the case where the golden span is empty while other span is predicted; \textbf{Under-extraction} refers to the case where the golden span is not empty while the predicted span is empty.

As shown in Figure~\ref{fig:error_analysis}, compared to the Single-EAE baseline PAIE, our DEEIA model reduces the number of errors from 312 to 292, indicating the effectiveness of DEEIA in capturing event correlations.
Compared to the baseline PAIE-multi, our DEEIA model reduces the number of errors from 358 to 292, especially decreasing the number of \textbf{Partial}, \textbf{Overlap}, and \textbf{Over-extract} errors. The ablation study shows that our proposed DE module mainly reduces the \textbf{Over-extract} and \textbf{Miss} errors, demonstrating that structural guidance can effectively help the model to deal with complex contexts. Meanwhile, the EIA module mainly reduces the \textbf{Miss} and \textbf{Partial} errors, indicating that this module offers event-specific contextual information for a better extraction of arguments.

\subsection{Comparison with Large Language Models}
\label{Sec:Comparison with Large Language Models}
Large language models (LLMs) have garnered substantial interest and attention from researchers, highlighting their extensive applicability across a wide array of tasks, such as text classification~\citep{chen-etal-2021-revisiting,luo-etal-2024-crosstune-black}, dialogue~\citep{zhang-etal-2023-xdial,zhang2024comprehensive}, offensive language detection~\citep{zhou-etal-2023-cultural, zhou-etal-2023-cross}, graph tasks~\cite{zeng2023substructure, zeng2022simple}, and in particular, the formation Extraction (IE) tasks~\cite{xu2023large, li2023semi, zhang2024ultra, zhou4703687rethinking, liu2022document, liu2024utilizing, xia2024enhancing, XIA2024110629}. In this paper, we  make a comparison with the recent state-of-the-art LLM-based approach presented in the work~\cite{zhou2023heuristics}, which utilizes LLMs for the EAE task. We report their experimental results in Table~\ref{tab:llm}.
The experiments are conducted on three prominent large language models: text-davinci-003~\cite{ouyang2022training}, gpt-3.5-turbo~\cite{ouyang2022training} and GPT-4~\cite{achiam2023gpt}.  These models are accessed via the public
APIs from OpenAI’s services\footnote{\url{https://openai.com/api/}}. 
As shown in Table~\ref{tab:llm}, compared to the supervised learning models, LLMs still show a significant performance gap in the EAE task. Additionally, the operational costs of large models are inherently high. Our approach outperforms LLMs in terms of efficiency, cost, and effectiveness in the document-level EAE task.
\begin{table}[tbp]
\centering
\setlength{\tabcolsep}{2.0mm}{
\begin{tabular}{lll}
\hline
\multirow{2}{*}{Method} & \multicolumn{2}{c}{RAMS} \\
                        & Arg-I       & Arg-C      \\ \hline
 HD-LoA~\cite{zhou2023heuristics}         \\               
\quad text-davinci-003        & 46.1       & 39.5      \\
\quad gpt-3.5-turbo           & 38.3       & 31.5       \\
\quad gpt-4                   & 50.4        & 42.8      \\ \hline
DEEIA (Ours)                  & 58.0        & 53.4       \\ \hline
\end{tabular}
}

\caption{Comparison with large language model method HD-LoA. We copy their experimental results.}
\label{tab:llm}
\end{table}

 \begin{table*}[htbp!]
\centering
\scriptsize
\begin{tabular}{@{ \quad}m{0.1\linewidth}<{\centering}m{0.25\linewidth}<{\centering}m{0.45\linewidth}<{\centering}@{}}
\toprule
\textbf{Dataset} & \textbf{Event Type} & \textbf{Natural Language Prompt} \\ \midrule
\multirow{20}{*}{ \textbf{WikiEvents}} & ArtifactExistence.\hspace{9em}DamageDestroyDisableDismantle.\hspace{9em}Damage & Damager (and Damager) damaged Artifact (and Artifact) using Instrument (and Instrument) in Place (and Place). \\
\specialrule{0em}{0pt}{2pt} \cline{2-3} \specialrule{0em}{2pt}{0pt}
& ArtifactExistence.\hspace{9em}DamageDestroyDisableDismantle.\hspace{9em}Destroy & Destroyer (and Destroyer) destroyed Artifact (and Artifact) using Instrument (and Instrument) in Place (and Place). \\
\specialrule{0em}{0pt}{2pt} \cline{2-3} \specialrule{0em}{2pt}{0pt}
& ArtifactExistence.\hspace{9em}DamageDestroyDisableDismantle.\hspace{9em}DisableDefuse & Disabler (and Disabler) disabled or defused Artifact (and Artifact) using Instrument (and Instrument) in Place (and Place). \\
\specialrule{0em}{0pt}{2pt} \cline{2-3} \specialrule{0em}{2pt}{0pt}
& ArtifactExistence.\hspace{9em}DamageDestroyDisableDismantle.\hspace{9em}Dismantle & Dismantler (and Dismantler) dismantled Artifact (and Artifact) using Instrument (and Instrument) from Components (and Components) in Place (and Place). \\
\specialrule{0em}{0pt}{2pt} \cline{2-3} \specialrule{0em}{2pt}{0pt}
& ArtifactExistence.\hspace{9em}DamageDestroyDisableDismantle.\hspace{9em}Unspecified & DamagerDestroyer (and DamagerDestroyer) damaged or destroyed Artifact (and Artifact) using Instrument (and Instrument) in Place (and Place). \\
\specialrule{0em}{0pt}{2pt} \cline{2-3} \specialrule{0em}{2pt}{0pt}
& ArtifactExistence.\hspace{9em}ManufactureAssemble.\hspace{9em}Unspecified & ManufacturerAssembler (and ManufacturerAssembler) manufactured or assembled or produced Artifact (and Artifact) from Components (and Components) using Instrument (and Instrument) at Place (and Place). \\
\specialrule{0em}{0pt}{2pt} \cline{2-3} \specialrule{0em}{2pt}{0pt}
& Cognitive.IdentifyCategorize.Unspecified & Identifier (and Identifier) identified IdentifiedObject (and IdentifiedObject) as IdentifiedRole (and IdentifiedRole) at Place (and Place). \\
\specialrule{0em}{0pt}{2pt} \cline{2-3} \specialrule{0em}{2pt}{0pt}
& Cognitive.Inspection.SensoryObserve & Observer (and Observer) observed ObservedEntity (and ObservedEntity) using Instrument (and Instrument) in Place (and Place). \\
\specialrule{0em}{0pt}{2pt} \hline \specialrule{0em}{2pt}{0pt}

\multirow{10}{*}{ \textbf{RAMS}} & artifactexistence.artifactfailure.\hspace{9em}mechanicalfailure & Mechanical artifact failed due to instrument at place. \\
\specialrule{0em}{0pt}{2pt} \cline{2-3} \specialrule{0em}{2pt}{0pt}
& artifactexistence.damagedestroy.n/a & DamagerDestroyer damaged or destroyed artifact using instrument in place. \\
\specialrule{0em}{0pt}{2pt} \cline{2-3} \specialrule{0em}{2pt}{0pt}
& artifactexistence.damagedestroy.damage & Damager damaged artifact using instrument in place. \\
\specialrule{0em}{0pt}{2pt} \cline{2-3} \specialrule{0em}{2pt}{0pt}
& artifactexistence.damagedestroy.destroy & Destroyer destroyed artifact using instrument in place. \\
\specialrule{0em}{0pt}{2pt} \cline{2-3} \specialrule{0em}{2pt}{0pt}
& artifactexistence.shortage.shortage & Experiencer experienced a shortage of supply at place. \\
\specialrule{0em}{0pt}{2pt} \cline{2-3} \specialrule{0em}{2pt}{0pt}
& conflict.attack.n/a & Attacker attacked target using instrument at place. \\
\specialrule{0em}{0pt}{2pt} \hline \specialrule{0em}{2pt}{0pt}

\multirow{10}{*}{ \textbf{MLEE}} & Cell\_proliferation & Cell proliferate or accumulate. \\
\specialrule{0em}{0pt}{2pt} \cline{2-3} \specialrule{0em}{2pt}{0pt}
& Development & Anatomical Entity develop or form. \\
\specialrule{0em}{0pt}{2pt} \cline{2-3} \specialrule{0em}{2pt}{0pt}
& Blood\_vessel\_development & Neovascularization or angiogenesis at Anatomical Location. \\
\specialrule{0em}{0pt}{2pt} \cline{2-3} \specialrule{0em}{2pt}{0pt}
& Growth & Growth of Anatomical Entity. \\
\specialrule{0em}{0pt}{2pt} \cline{2-3} \specialrule{0em}{2pt}{0pt}
& Death & Death of Anatomical Entity. \\
\specialrule{0em}{0pt}{2pt} \cline{2-3} \specialrule{0em}{2pt}{0pt}
& Breakdown & Anatomical Entity degraded or damaged. \\
\specialrule{0em}{0pt}{2pt} \cline{2-3} \specialrule{0em}{2pt}{0pt}
& Remodeling & Tissue remodeling or changes. \\
\specialrule{0em}{0pt}{2pt} \cline{2-3} \specialrule{0em}{2pt}{0pt}
& Synthesis & Synthesis of Drug/Compound. \\
\bottomrule
\end{tabular}
\caption{Example of Prompts in Tabular Format}
\label{tab:prompt}

\end{table*}

\end{document}